\documentclass[3p,times]{elsarticle}

\usepackage{ecrc}

\volume{00}

\firstpage{1}

\journalname{Procedia Computer Science}

\runauth{}

\jid{procs}

\jnltitlelogo{Procedia Computer Science}

\CopyrightLine{2011}{Published by Elsevier Ltd.}

\usepackage{amssymb}

\usepackage[figuresright]{rotating}

\usepackage[utf8]{inputenc}
\usepackage{microtype}
\usepackage[T1]{fontenc}    
\usepackage{url}            
\usepackage{booktabs}       
\usepackage{amsfonts}       
\usepackage{nicefrac}       
\usepackage{amsthm}
\usepackage{algorithm}
\usepackage{algorithmic}
\usepackage{graphicx}
\usepackage[percent]{overpic}
\usepackage{booktabs} 
\usepackage{bm}
\usepackage{hyperref}
\usepackage{amsmath}
\usepackage{amssymb}
\usepackage{cleveref}
\usepackage{enumitem}
\usepackage{amssymb}
\usepackage{mathtools}
\newtheorem{exmp}{Example}[section]

\usepackage{multirow}
\usepackage[normalem]{ulem}
\useunder{\uline}{\ul}{}
\usepackage{caption}
\usepackage{subcaption}

\usepackage{afterpage}
\usepackage{makecell}
\usepackage{tabulary}
\usepackage{xcolor}
\definecolor{revisionpurple}{RGB}{112,48,160}
\colorlet{purple}{revisionpurple}
\usepackage{colortbl}
\usepackage{prettyref}
\usepackage{framed}
\newcommand{\pref}[1]{\prettyref{#1}}

\newcommand{\savehyperref}[2]{\texorpdfstring{\hyperref[#1]{#2}}{#2}}
\newrefformat{eq}{\savehyperref{#1}{Eq. \textup{(\ref*{#1})}}}
\newrefformat{eqn}{\savehyperref{#1}{Eq.~\textup{(\ref*{#1})}}}
\newrefformat{lem}{\savehyperref{#1}{Lemma~\ref*{#1}}}
\newrefformat{assump}{\savehyperref{#1}{Assumption~\ref*{#1}}}
\newrefformat{defi}{\savehyperref{#1}{Definition~\ref*{#1}}}\newrefformat{tab}{\savehyperref{#1}{Table~\ref*{#1}}}
\newrefformat{lemma}{\savehyperref{#1}{Lemma~\ref*{#1}}}
\newrefformat{line}{\savehyperref{#1}{Line~\ref*{#1}}}
\newrefformat{thm}{\savehyperref{#1}{Theorem~\ref*{#1}}}
\newrefformat{corr}{\savehyperref{#1}{Corollary~\ref*{#1}}}
\newrefformat{cor}{\savehyperref{#1}{Corollary~\ref*{#1}}}
\newrefformat{sec}{\savehyperref{#1}{Section~\ref*{#1}}}
\newrefformat{app}{\savehyperref{#1}{\textbf{Appendix}~\ref*{#1}}}
\newrefformat{ex}{\savehyperref{#1}{Example~\ref*{#1}}}
\newrefformat{fig}{\savehyperref{#1}{Figure~\ref*{#1}}}
\newrefformat{alg}{\savehyperref{#1}{Algorithm~\ref*{#1}}}
\newrefformat{rem}{\savehyperref{#1}{Remark~\ref*{#1}}}
\newrefformat{conj}{\savehyperref{#1}{Conjecture~\ref*{#1}}}
\newrefformat{prop}{\savehyperref{#1}{Proposition~\ref*{#1}}}
\newrefformat{proto}{\savehyperref{#1}{Protocol~\ref*{#1}}}
\newrefformat{prob}{\savehyperref{#1}{Problem~\ref*{#1}}}
\newrefformat{claim}{\savehyperref{#1}{Claim~\ref*{#1}}}
\newrefformat{que}{\savehyperref{#1}{Question~\ref*{#1}}}
\newrefformat{op}{\savehyperref{#1}{Open Problem~\ref*{#1}}}
\newrefformat{fn}{\savehyperref{#1}{Footnote~\ref*{#1}}}
\newrefformat{property}{\savehyperref{#1}{Property~\ref*{#1}}}

\usepackage{xspace}
\usepackage{amsmath}
\usepackage{amsthm}
\usepackage{bbm}
\usepackage{algorithm}
\usepackage[algo2e, vlined, ruled]{algorithm2e}
\usepackage{algorithmic}
\usepackage{mathrsfs}
\usepackage{bm}
\usepackage{makecell}
\usepackage{tabulary}

\newcommand{\calD}{\mathcal{D}}

\newcommand{\calL}{\mathcal{L}}

\newcommand{\calB}{\mathcal{B}}

\newcommand{\E}{\mathbb{E}}

\newcommand{\red}[1]{{\color{black}#1}}
\newcommand{\currentrev}[1]{{\color{red}#1}}
\newcommand{\bluerev}[1]{{\color{black}#1}}
\newcommand{\yellow}[1]{\red{#1}}

\newtheorem*{theorem*}{Theorem}

\newtheorem{remark}{Remark}
\newtheorem{assumption}{Assumption}[section]

\newtheorem*{lemma*}{Lemma}
\newtheorem{definition}{Definition}[section]

\theoremstyle{definition}

\newtheorem{thm}{Theorem}[section]

\newtheorem{lem}[thm]{Lemma}
\newtheorem{cor}{Corollary}[section]

\newrefformat{exmp}{\savehyperref{#1}{Example~\ref*{#1}}}

\hypersetup{hidelinks}

\usepackage{xcolor}

\colorlet{red}{black}
\begin{document}

\begin{frontmatter}



\renewcommand{\thefootnote}{\fnsymbol{footnote}}

\dochead{}

\title{Theoretically Principled Federated Learning for Balancing Privacy and Utility}


\author[label1]{Xiaojin Zhang
\footnote{National Engineering Research Center for Big Data Technology and System, Services Computing Technology and System Lab, Cluster and Grid Computing Lab, School of Computer Science and Technology, Huazhong University of Science and Technology, Wuhan, 430074, China}}
\address[label1]{Huazhong University of Science and Technology, China}
\author[label2]{Wenjie Li}
\author[label1]{Yiming Li}
\author[label1]{Wei Chen}
\address[label2]{Tsinghua University, China}
\author[label2,label4]{Shu-Tao Xia}
\address[label4]{Peng Cheng Laboratory, China}
\author[label5,label6]{Qiang Yang\footnote{Corresponding Author}}
\address[label5]{WeBank, China}
\address[label6]{The Hong Kong Polytechnic University, China}




\begin{abstract}
      Federated learning aims to enable collaborative model training while preserving data privacy. Distortion-based protection mechanisms are commonly employed to achieve this objective. However, the adoption of distortion can lead to a degradation in performance. Existing research predominantly focuses on privacy or utility independently, lacking a unified approach. In this work, we propose a unified learning algorithm for balancing the trade-off between privacy and utility in federated learning. Our algorithm is specifically designed for distortion-based privacy protection mechanisms and \bluerev{can be applied to privacy measurements that map distortion to a scalar leakage value}. By adaptively adjusting the distortion for each model parameter based on its impact on model utility, our learning algorithm achieves a superior utility-privacy trade-off. \bluerev{Under the stated regularity conditions, we analyze the utility performance of the distortion-learning procedure and the convergence behavior of the resulting perturbed federated optimization.} \bluerev{Experiments on MNIST and Fashion-MNIST} provide empirical evidence supporting the superior utility performance of our method compared to baseline methods, while maintaining the same privacy budget settings.
\end{abstract}


\end{frontmatter}



\section{Introduction}

The proliferation of extensive datasets has created a demand for distributed learning techniques. Federated learning (FL) \cite{mcmahan2016federated,mcmahan2017communication,konevcny2016federated,konevcny2016federated_new,he2024reinforcement} has emerged as a promising paradigm that enables multiple participants to collaboratively train a machine learning model while preserving the privacy of their sensitive data. FL has made notable advancements in privacy-preserving machine learning systems, specifically in addressing the challenges posed by "semi-honest" adversaries. These adversaries strictly adhere to the principles of the federated learning framework but aim to infer sensitive information of other participants based on the exposed model parameters \cite{zhao2022accel,li2024model}. The degree of statistical dependence between private data and publicly transmitted information determines the extent to which a semi-honest entity can deduce private data from the exchanged information. Recent research has revealed that semi-honest adversaries can exploit the gradients of trained models to reconstruct private training images with high fidelity at the pixel level, a phenomenon known as gradient leakage attacks. Notable examples of such attacks include DLG \cite{zhu2020deep}, Inverting Gradients \cite{geiping2020inverting}, Improved DLG \cite{zhao2020idlg}, and GradInversion \cite{yin2021see}. In this context, this paper aims to introduce a theoretically principled approach to federated learning that carefully balances the trade-off between privacy and utility.

Early attempts to address privacy attacks have involved various techniques, including homomorphic encryption (HE) \cite{hardy2017private}, differential privacy (DP) \cite{abadi2016deep}, and gradient compression (GC) \cite{lin2018deep}. HE and MPC offer protection for private data without compromising model performance, but they often introduce significant computation and communication overhead, particularly for deep neural networks. Moreover, these methods do not ensure the security of clients' private data after decryption during server aggregation \cite{lam21b}. DP and GC protect data privacy by introducing \textit{distortion}, such as adding noise or compressing shared model updates, which can lead to a substantial degradation in model performance. To achieve a balance between privacy and performance, recent works \cite{wei2021gradient,noble22a,zhu2021fine,shen2022performance} have explored approaches that leverage fine-grained DP or regularization to mitigate the impact of noise on model performance. However, these protection mechanisms still require manual tuning of protection hyperparameters, necessitating significant engineering efforts. 

In contrast, the primary contribution of this work is a novel protection mechanism that learns the distortion jointly with the optimization of model parameters. We propose an adaptive learning algorithm designed to automatically attain a desirable balance between privacy and utility. This algorithm dynamically adjusts the distortion for each model parameter based on its specific impact on model performance. Furthermore, \textcolor{black}{under explicit regularity conditions, we establish an average first-order stationarity guarantee for the practical nonconvex $L_2$-annular projected-SGD updates used by the PL branch, \bluerev{derive the $O(1/M)$ global utility-loss gap in the convex exact-gradient regime}, and characterize the average stationarity of perturbed FedSGD.} This shift from 'static configuration' to 'dynamic learning' offers a new, theoretically principled approach to solving the fundamental privacy-utility trade-off in federated learning.

Our adaptive distortion-learning framework is designed to address the limitations of traditional static distortion strategies. The static approach, where a uniform distortion is applied for privacy, is advantageous for its simplicity and efficiency. However, this simplicity comes at the cost of flexibility. Thestatic approach fails to consider client-specific heterogeneity, such as differences in data distributions or privacy requirements. In contrast, our framework adaptively personalizes the distortion for each client, achieving a superior balance by accommodating their unique data characteristics and privacy needs. The primary advantage of this framework lies in this fine-grained optimization and flexibility. Its main drawback is the increased computational and management complexity required to enable this personalization, which is worthwhile for achieving a reasonable trade-off between privacy and utility in diverse federated learning environments.

\paragraph{\textbf{Our Contributions}}
First, we formulate the balance between privacy and utility as an optimization problem with respect to distortion variables. Our objective is to identify appropriate distortion values that minimize the loss in utility while ensuring that the potential privacy leakage remains below an acceptable threshold.
\begin{itemize}[leftmargin=*]
   \item Regarding privacy, we build upon existing privacy measurements such as information leakage \cite{du2012privacy, zhang2024deciphering} and Kullback-Leibler divergence \cite{eilat2021bayesian}. In the context of federated learning, previous studies \cite{zhang2022no, zhang2023trading} introduced a privacy measurement based on Jensen--Shannon divergence, which quantifies privacy loss by comparing prior and posterior beliefs. However, these approaches assume prior knowledge of the distribution for private information, which is often impractical in real-world scenarios where raw data distributions are complex and non-trivial. In contrast, our proposed privacy measurement does not rely on prior distribution knowledge and is valuable for designing algorithms with theoretical guarantees.
   \item Concerning utility, we depart from previous approaches that solely consider the expected distance between the original model and the perturbed model \cite{du2012privacy}. Instead, our method evaluates utility loss by considering the performance of both the original model and the perturbed model. This direct assessment of the model's utility adds complexity to the problem, presenting a more challenging optimization task. 
   \item \textcolor{black}{To evaluate the effectiveness of the proposed protection mechanisms, we draw inspiration from the concept of regret commonly used in learning theory \cite{slivkins2019introduction, lattimore2020bandit}. For the nonconvex $L_2$-annular PL branch, we evaluate the projected-SGD trajectory through first-order stationarity; when the distortion subproblem and feasible set are convex and exact gradients are used, we additionally quantify its utility-loss gap to an optimal distortion.}
\end{itemize}
\textcolor{black}{Second, we propose an algorithmic framework aimed at facilitating the balance between utility and privacy in protection mechanisms. Our approach learns parameter-wise distortions to reduce utility loss while enforcing privacy constraints. To accomplish this, we use stochastic-gradient updates followed by an intensity-feasibility operation; this operation is an exact Euclidean projection for the $L_2$-annular PL branch analyzed below.}
\begin{itemize}[leftmargin=*]
   \item \textcolor{black}{To address privacy constraints, we propose a learning-based protection mechanism that seeks to reduce utility loss over the feasible distortion region. This mechanism allows for fine-tuned balancing between privacy and utility in privacy-preserving federated learning. Rather than applying a standardized approach of adding the minimal possible noise to meet the privacy constraint, we adapt the parameter-wise distortion while respecting its prescribed intensity bounds.}
   \item \textcolor{black}{For the nonconvex $L_2$-annular projected-SGD realization used by the practical PL learner, we establish an $O(1/M)$ transient rate to a first-order stationary neighborhood without requiring an unbiased implementation gradient (\pref{thm: distortion_stationarity_mt}). Under the additional convex-set, convex-objective, and exact-gradient conditions, \bluerev{an $O(1/M)$ utility-loss gap to the optimum is established as a convex-case corollary} (\pref{cor: convex_suboptimality_mt}).} As a comparison, \cite{rassouli2019optimal} formulated the balance between privacy and utility as optimization problems beyond the field of federated learning. However, the optimization problems might not be tractable in practice and they only provided a closed-form solution for some Boolean-valued special cases.
   \item \yellow{We provide a convergence analysis for the perturbed FedSGD updates in our proposed framework under standard smoothness and stochastic-gradient assumptions. Additionally, we establish that the time-averaged expected squared gradient norm is bounded by a stationary neighborhood whose size depends on the stochastic-gradient variance and the expected squared norm of the aggregate distortion (\pref{thm: converge_rate_mt}).}
   \item We apply our algorithm framework to various privacy measurements, \textcolor{black}{including a Laplace-based distortion-to-leakage measurement}, as well as a novel measurement that evaluates the amount of privacy leakage by examining the gap between estimated and true data. Empirical results obtained from benchmark datasets provide evidence that our method outperforms baseline methods in terms of utility \textcolor{black}{at fixed budgets}.
\end{itemize}


\section{Related Work}



\paragraph{\textbf{Attacking and Defense Mechanisms in Federated Learning}}\label{sec:related:attack}

In the field of HFL, several studies (\cite{zhu2019dlg,zhu2020deep,geiping2020inverting,zhao2020idlg,yin2021see}) showcase how adversaries can leverage the transmitted gradients or model information to accurately reconstruct private data information. To protect clients' private data, various general protection mechanisms have been proposed, including Homomorphic Encryption (HE) \cite{gentry2009fully,batchCryp}, Differential Privacy (DP) \cite{geyer2017differentially,truex2020ldp,abadi2016deep}, and Gradient Compression (GC) \cite{nori2021fast}. However, these mechanisms have their limitations. HE and MPC incur significant computation and communication overheads, while DP and GC can cause a degradation in model performance.
Consequently, considerable efforts have been devoted to preserving data privacy while maintaining satisfactory model performance. For instance, some approaches (\cite{wei2021gradient,noble22a}) employ record-level client DP, while others (\cite{zhu2021fine}) allocate layer-wise noise to conserve privacy budget. Another approach proposed by \cite{shen2022performance} leverages a perturbation regularizer to mitigate the impact of noise on model performance. Additionally, \cite{wu2022fedcg} utilizes split learning to protect privacy and employs a generative neural network to compensate for performance loss.

\paragraph{\textbf{Balance between Privacy and Utility}} 
The optimal balance between privacy and utility in federated learning can be mathematically expressed as a constrained optimization problem. The objective is to minimize the loss in utility while adhering to a predefined constraint on privacy leakage \cite{zhang2023towards, zhang2023trading,zhang2023meta}. Previous studies have explored different aspects of balance between privacy and utility.
For instance, \cite{duchi2013local} demonstrated balance between privacy and convergence rate in locally private settings. \cite{rassouli2019optimal} demonstrated that the optimal trade-off can be addressed through linear programming. However, they only furnished a closed-form solution exclusively applicable to a particular binary scenario.
Asymptotic results on the rate-distortion-equivocation region with an increasing number of sampled data were investigated by \cite{reed1973information, yamamoto1983source, sankar2013utility}. \cite{sankar2013utility} quantified utility using accuracy and privacy using entropy. By leveraging rate-distortion theory, they defined a utility-privacy balance region for independent and identically distributed datasets. However, extending the quantification of utility-privacy trade-offs to general scenarios still presents a challenging open problem.
In a unified privacy-distortion framework, distortion was measured by computing the expected Hamming distance between the inputs and outputs \cite{wang2016relation}. Additionally, they evaluated privacy leakage using DP, identifiability, and mutual-information as separate measurements. They also established the relationship between these different privacy metrics. 

\paragraph{\textbf{Positioning Our Work in a Broader Context}} 
It is worth noting that the principle of balancing competing objectives is a cornerstone of many information security and machine learning tasks. For instance, in the domain of image encryption, works such as \cite{feng2024exploiting} leverage hyperchaotic maps to generate noise-like key streams for securing image data, aiming to maximize security (confidentiality) while maintaining computational efficiency. Although their approach of direct data encryption for confidentiality differs from our method of adding distortion to model parameters for privacy, both paradigms rely on the fundamental properties of chaotic or random noise to achieve their respective protection goals. Furthermore, the challenge of managing the impact of "noise" on model performance is a recurring theme in machine learning, albeit with different interpretations of "noise". In our context, we strategically inject noise to preserve privacy. Conversely, in fields like image captioning, the goal is often to filter out inherent noise, which corresponds to irrelevant features. For example, the SAMT-generator proposed by \cite{yang2024samt} introduces a "second-attention" mechanism to identify and suppress irrelevant visual information, thereby enhancing the model's focus on salient objects and improving caption quality. Interestingly, this mirrors a core principle of our work in an abstract sense: just as their attention mechanism identifies the importance of different visual features, our algorithm assesses the impact of different model parameters on utility to apply distortion adaptively. This shared concept of importance-aware adjustments highlights a convergent line of thinking across different machine learning applications.

\section{Preliminaries}
In this work, our focus is on the HFL setting. We consider a scenario with a total of $K$ clients, and we denote $\calD^{(k)}$ as the dataset owned by client $k$. The goal of the $K$ clients is to collectively train a global model
    \begin{align*}
        W^* &= \arg\min_{W}\sum_{k = 1}^K \frac{n^{(k)}}{n}\ \calL^{(k)}(W),
    \end{align*} 
where $n^{(k)}$ denotes the size of the dataset $\calD^{(k)}$, $n=\sum_{k=1}^K n^{(k)}$, and $\mathcal{L}^{(k)}(W)$ represents the loss of predictions made by the model parameter $W$ on dataset $\calD^{(k)}$.



\paragraph{\textbf{Threat Model}} In our study, we assume that the server acts as a \textit{semi-honest} adversary. This means that the server faithfully follows the federated training protocol but has the capability to mount gradient inversion attacks, as demonstrated in \cite{zhu2019dlg}, where the server can reconstruct the private data of participating clients by observing the shared model updates.




{\color{red}
\paragraph{\textbf{Training Framework}} We follow the model update procedure of \textit{federated SGD} (FedSGD) (\cite{mcmahan2017communication}). Let $p^{(k)}=n^{(k)}/n$ denote the aggregation weight of client $k$. At communication round $t$, the privacy-preserving training procedure is:
\begin{itemize}[leftmargin=*]
    \item after receiving $W_{t-1}$, client $k$ computes the undistorted local update $\breve W_t^{(k)}=W_{t-1}-\eta\nabla\calL^{(k)}(W_{t-1})$;
    \item client $k$ selects a parameter-wise distortion $\alpha_t^{(k)}$ and releases $W_t^{(k)}=\breve W_t^{(k)}+\alpha_t^{(k)}$;
    \item the server aggregates $W_t=\sum_{k=1}^Kp^{(k)}W_t^{(k)}$ and broadcasts $W_t$.
\end{itemize}
\textcolor{black}{Let $\breve{\calB}_t^{(k)}$ denote the actual data collection (the full local dataset or the realized mini-batch, as applicable) whose empirical-average loss is used to form $\nabla\calL^{(k)}(W_{t-1})$. With the per-example exposed-update map introduced in \bluerev{\ref{sec: discuss_measurements_for_privacy}}, its aggregate satisfies $\overline g_t^{(k)}(\breve{\calB}_t^{(k)})=\breve W_t^{(k)}$ directly from this FedSGD update.}
Here, $W_t$ is the protected aggregate model and $\eta$ is the local learning rate. The tensor collection $\alpha_t^{(k)}$ has the same shape as the model parameters. Its scalar intensity is denoted by $\beta_t^{(k)}=h(\alpha_t^{(k)})$, where $h$ is specified by the protection mechanism. We reserve
\begin{align}
    \Delta_t=\sum_{k=1}^Kp^{(k)}\alpha_t^{(k)}
    \label{eq: aggregate_distortion}
\end{align}
for the aggregate distortion appearing in the server update; $\Delta_t$ is therefore a collection of tensors, not a scalar norm.

We call a protection mechanism \textit{distribution-dependent} when its privacy claim requires $\alpha_t^{(k)}$ to follow a prescribed probability law, as in the Laplace or Gaussian mechanism. A \textit{distribution-independent} mechanism constrains a property of the realized distortion, such as $h(\alpha_t^{(k)})$, without requiring a particular law. These two cases lead to different feasible optimization variables and are distinguished below.
}

\begin{table*}[!ht]
\footnotesize
  \centering
  \setlength{\belowcaptionskip}{15pt}
  \caption{Notations.} 
  \label{table: notation}
    \begin{tabular}{ccc}
    \toprule
    Notation & Meaning & Value Type\cr
    \midrule
    $\epsilon_p^{(k)}$ & privacy leakage of client $k$ & scalar\cr
    \red{$\epsilon_{u,t}^{(k)}$} & \red{signed utility loss of client $k$ at round $t$} & \red{scalar}\cr
    $\calL$ & loss function & scalar\cr
    \red{$W_t^{(k)},W_t$} & \red{protected local and aggregate model parameters} & \red{collections of tensors}\cr
    \red{$\breve W_t^{(k)},\breve W_t$} & \red{undistorted local and aggregate model parameters} & \red{collections of tensors}\cr
    \midrule
    \currentrev{$\alpha_t^{(k)}$} & \currentrev{parameter-wise distortion applied by client $k$} & \currentrev{collection of tensors}\cr
    \currentrev{$\beta_t^{(k)}$} & \currentrev{scalar distortion intensity $h(\alpha_t^{(k)})$} & \currentrev{scalar}\cr
    \currentrev{$\Delta_t$} & \currentrev{aggregate distortion $\sum_kp^{(k)}\alpha_t^{(k)}$} & \currentrev{collection of tensors}\cr
    \textcolor{black}{$\chi_t^{(k)}$} & \textcolor{black}{privacy budget of client $k$ at round $t$} & scalar\cr
    \bottomrule
    \end{tabular}
\end{table*}

\textcolor{black}{Notation convention: a client index is written as a superscript $(k)$ and a communication-round index as a subscript $t$. These indices are suppressed only when a statement is explicitly made for a generic fixed client or round.}


\section{The Framework}

\begin{figure}
    \centering
    \includegraphics[width=0.8\textwidth]{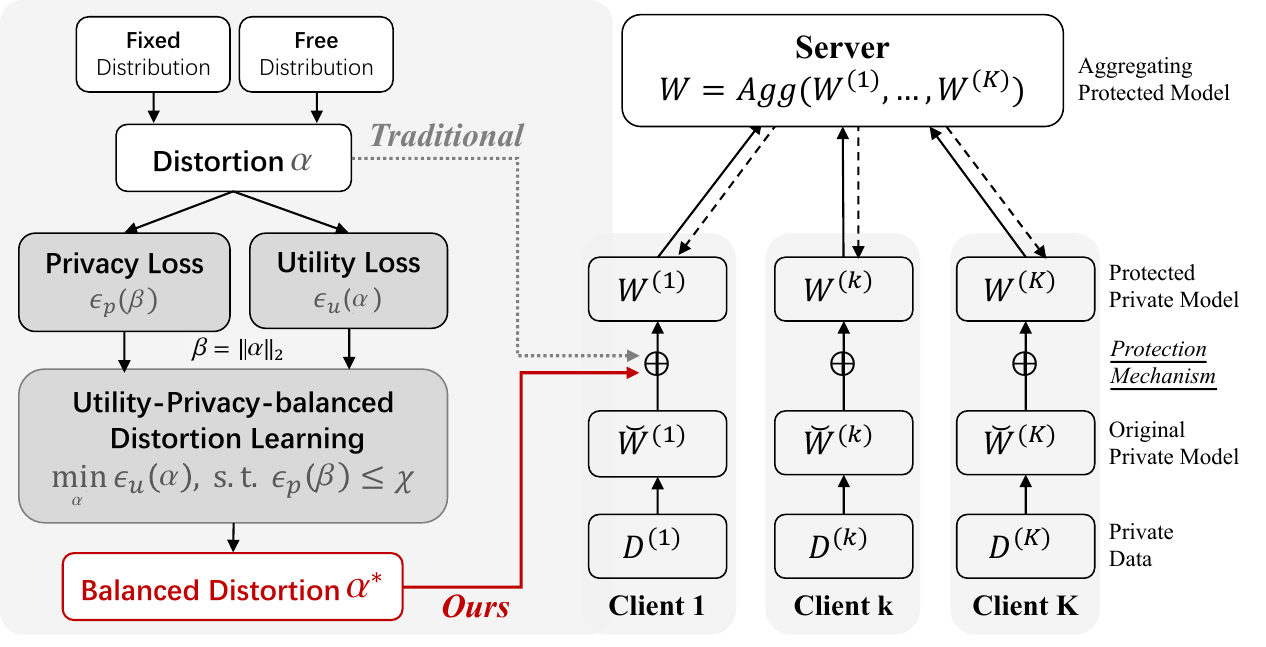}
    \caption{An overview of our balanced framework for utility and privacy protection, with the first client serving as an illustrative example showing our detailed protection schema. Each client independently conducts the distortion process.}
    \label{fig:nfl_overview}
\end{figure}

{\color{black}
In this work, we learn $\alpha_t^{(k)}$ adaptively to reduce the utility cost of protection while satisfying a prescribed privacy constraint. We first define the utility objective and the privacy constraint using the notation above.

\paragraph{\textbf{Utility Loss}}
The undistorted aggregate at round $t$ is
\begin{align}
    \breve W_t
    =\sum_{k=1}^Kp^{(k)}\breve W_t^{(k)}
    =W_{t-1}-\eta\sum_{k=1}^Kp^{(k)}\nabla\calL^{(k)}(W_{t-1}).
\end{align}
{\color{red}
The undistorted local update $\breve W_t^{(k)}$ is computed from the received model $W_{t-1}$ without applying the current-round distortion $\alpha_t^{(k)}$. Accordingly, the undistorted aggregate is $\breve W_t=\sum_{k=1}^Kp^{(k)}\breve W_t^{(k)}$. Importantly, $W_{t-1}$ may already contain the accumulated effects of distortions applied in previous rounds. Therefore, $\breve W_t^{(k)}$ and $\breve W_t$ exclude only the distortion newly introduced at round $t$; they do not represent a fully distortion-free training trajectory. This construction isolates the marginal utility effect of the current-round protection.

The protected aggregate is consequently
\begin{align}\label{eq: distorted_model_parameter_mt}
    W_t
    =\sum_{k=1}^Kp^{(k)}\left(\breve W_t^{(k)}+\alpha_t^{(k)}\right)
    =\breve W_t+\Delta_t.
\end{align}
The utility loss evaluated for client $k$ at round $t$ is defined as the signed change in its loss caused by the current-round protection relative to the corresponding undistorted local model:
\begin{align}\label{eq: utility_loss_definition}
\epsilon_{u,t}^{(k)}
&=
\mathcal L^{(k)}(W_t^{(k)})
-
\mathcal L^{(k)}(\breve W_t^{(k)}).
\end{align}
This sign convention is used throughout: a positive value means that protection increases training loss, whereas a negative value means that the distortion decreases it.
}

\textbf{Remark:}
The utility loss of the federated system is the average utility loss over rounds and clients, $\epsilon_u=\frac{1}{K}\frac{1}{T}\sum_{k=1}^K\sum_{t=1}^T\epsilon_{u,t}^{(k)}$. We can use an alternative metric, which is defined as $\epsilon_u^{\mathrm{final}}=\frac{1}{K}\sum_{k=1}^K\epsilon_{u,T}^{(k)}$. This metric focuses on the last iteration and takes the average over $K$ clients. \textcolor{black}{The global utility-loss gap in \pref{cor: convex_suboptimality_mt} can be averaged over the corresponding clients and rounds only when its convexity and exact-gradient conditions hold for each included subproblem. The general nonconvex guarantee in \pref{thm: distortion_stationarity_mt} instead concerns the average first-order stationarity of each inner distortion-optimization trajectory.}
}

\paragraph{\textbf{Privacy Leakage}}
In this work, we \textcolor{black}{discuss} two measurements to quantify privacy leakage. The first measurement is differential privacy, which is a commonly used privacy metric \red{defined for a randomized release mechanism and therefore depends on the prescribed distribution of $\alpha_t^{(k)}$}. We refer readers to Section III of \cite{zhu2021fine} for more details on this measurement. \currentrev{For the adaptive realized-distortion algorithm and the experiments, the Laplace form is used as a scale-to-budget calibration; the implemented learner is not claimed to provide a formal differential-privacy release.}

\textcolor{black}{Now, we introduce the second measurement in detail. The semi-honest attacker utilizes an optimization algorithm \cite{zhu2020deep, geiping2020inverting, zhao2020idlg, yin2021see} to infer the original data by matching the transmitted protected model information $W$. For a fixed round and client, use the shorthand $\alpha^{(k)}=\alpha_t^{(k)}$, $\beta^{(k)}=\beta_t^{(k)}$, $W^{(k)}=W_t^{(k)}$, and $\breve W^{(k)}=\breve W_t^{(k)}$. Let $\beta^{(k)}=h(\alpha^{(k)})$ denote the scalar intensity of the parameter-wise distortion applied by client $k$. For the attack-specific score below, we set $h(\alpha)=\|\alpha\|_2$. Because the algorithm releases $W^{(k)}=\breve W^{(k)}+\alpha^{(k)}$, this gives $\beta^{(k)}=\|\alpha^{(k)}\|_2=\|W^{(k)}-\breve W^{(k)}\|_2$ by construction. The privacy leakage is assessed from the discrepancy between the attacker's reconstruction and the actual private data. Fixing $I$ (the total number of attack iterations), we use the following attack-specific leakage score:}

{\color{black}
\begin{align}\label{eq: measurement_for_privacy}
    \epsilon_p^{(k)}(\beta^{(k)})
    =1-\frac{c_a\beta^{(k)}+c_2c_bI^{p-1}}{4D}.
\end{align}
}
\textcolor{black}{Here, $0<c_a\le c_b$ are the constants in the bi-Lipschitz condition relating distances in the data space to distances in the per-example exposed-update space. For fixed $W_{t-1}$ and $\eta$, that map is an affine rescaling of the gradient map, and the scaling is absorbed into the constants. For the selected attack horizon $I$, the positive constant $c_2$ and exponent $p\in(0,1)$ specify an upper envelope $c_2I^p$ for the attacker's cumulative aggregate matching residual with respect to the actually released update $W$; they do not refer to a hypothetical dataset whose update must equal $W$. If multiple horizons are considered, the same constants are required over those horizons. The factor $c_b$ enters through the conservative relaxation $c_a\le c_b$ in \bluerev{\ref{sec: discuss_measurements_for_privacy}}. Detailed assumptions and the direct derivation are given in \bluerev{\pref{assump: two_sided_Lipschitz}}, \bluerev{\pref{assump: matching_residual}}, and \bluerev{\pref{lem: bound_for_privacy_leakage}}. The constant $D$ bounds the distance between a reconstruction and the corresponding private example. Thus, increasing the realized distortion intensity decreases this score. Because the score is conditional on a fixed gradient-inversion attack protocol and its matching-residual envelope, it is an attack-specific leakage surrogate rather than a mechanism-independent privacy guarantee.}










\paragraph{\textbf{The Optimization Problem}}

Let $\epsilon_p^{(k)}$ represent the privacy leakage associated with client $k$. \currentrev{Let $\mathcal D_\alpha\subseteq\mathbb R^d$ denote the distortion space with the same dimension as the vectorized model parameters, and let $\mathfrak P\subseteq\mathcal P(\mathcal D_\alpha)$ be a feasible family of noise distributions (e.g., a parametric Laplace/Gaussian family).} \currentrev{For a candidate distortion $\alpha$, define
\begin{align}
\epsilon_{u,t}^{(k)}(\alpha)
&=\mathcal L^{(k)}(\breve W_t^{(k)}+\alpha)
-\mathcal L^{(k)}(\breve W_t^{(k)}),
\label{eq: candidate_utility_loss}
\end{align}
so that the utility loss in \pref{eq: utility_loss_definition} is obtained by evaluating this function at $\alpha=\alpha_t^{(k)}$.} Previous work \cite{zhang2022no} formulates the balance between utility and privacy in federated learning by selecting a release distribution $P_t^{(k)}\in\mathfrak P$. The objective is to minimize the expected utility loss $\epsilon_{u,t}^{(k)}(P_t^{(k)})\triangleq\E_{\alpha\sim P_t^{(k)}}[\epsilon_{u,t}^{(k)}(\alpha)]$ across rounds and clients subject to a predefined constraint on privacy leakage:

\begin{align} \label{eq: constraint_optimization_problem_ul}
\begin{array}{r@{\quad}l@{}l@{\quad}l}
\quad\min\limits_{\{P_t^{(k)}\}_{t\in [T], k\in [K]}}& \frac{1}{K}\frac{1}{T}\sum_{k=1}^K\sum_{t=1}^T\red{\epsilon_{u,t}^{(k)}}(P_t^{(k)}),\\
\text{s.t.,} & \epsilon_{p}^{(k)}(P_t^{(k)})\le\chi_t^{(k)}, \forall t\in [T], \forall k\in [K].
\end{array}
\end{align}
However, the requirement that $P_t^{(k)}$ belong to a \red{prescribed feasible distribution family} limits the flexibility of applying arbitrary distortions and dynamically adjusting their intensity. To relax this restriction, \red{at each round $t$ and conditioned on $W_{t-1}$,} we define the following optimization problem for a distribution-independent privacy measurement:
\begin{align} \label{eq: optimization_problem_for_wide}
\begin{array}{r@{\quad}l@{}l@{\quad}l}
\quad\min\limits_{\red{\{\alpha_t^{(k)}\}_{k\in [K]}}}& \frac{1}{K}\sum_{k=1}^K\red{\epsilon_{u,t}^{(k)}}(\alpha_t^{(k)}),\\
\text{s.t.,} & \epsilon_{p}^{(k)}(\beta_t^{(k)})\le\chi_t^{(k)}, \red{\forall k\in [K]}.\\
\quad\quad\quad &\beta_t^{(k)} = h(\alpha_t^{(k)}), \red{\forall k\in [K]}.
\end{array}
\end{align}


\currentrev{Here, $\alpha_t^{(k)}$ and $\beta_t^{(k)}$ represent an arbitrary parameter-wise distortion and its corresponding scalar intensity for client $k$, respectively, where $\beta_t^{(k)}=h(\alpha_t^{(k)})$. The function $h$ maps a distortion value, which may be a vector or a collection of tensors, to a scalar representing its overall intensity. For example, if $\alpha_t^{(k)}$ is arbitrarily generated noise, $h$ can be the $L_2$-norm and $\beta_t^{(k)}$ is its total norm. The privacy-leakage function $\epsilon_p^{(k)}$ quantifies the leakage associated with $\beta_t^{(k)}$ and is required to remain below the budget $\chi_t^{(k)}$. This realized-distortion formulation applies directly to a distribution-independent protection mechanism.}

\currentrev{For a monotonically decreasing leakage score, its algebraic inverse gives the budget-induced lower bound. We use $g^{(k)}$ more generally as a conservative budget-to-intensity map that also enforces any condition required for the score bound. Hence, $\beta_t^{(k)}\ge g^{(k)}(\chi_t^{(k)})$ is sufficient under those conditions; for \pref{eq: measurement_for_privacy}, $g^{(k)}$ is the max-form in the following remark and Example~\ref{exmp: distribution_free_prot_mechanism}, not the unconstrained inverse alone.} \red{With the user-specified upper bound $u_t^{(k)}$, the per-round problem used by our algorithm is as follows:}

\begin{align}\label{eq: optimization_problem_for_lemma}
\begin{array}{r@{\quad}l@{}l@{\quad}l}
\quad\min\limits_{\red{\{\alpha_t^{(k)}\}_{k\in [K]}}}& \frac{1}{K}\sum_{k=1}^K\red{\epsilon_{u,t}^{(k)}}(\alpha_t^{(k)}),\\
\text{s.t.,} & g^{(k)} (\chi_t^{(k)})\le \beta_t^{(k)}\le \red{u_t^{(k)}}, \red{\forall k\in [K]}.\\
\quad\quad\quad &\beta_t^{(k)} = h(\alpha_t^{(k)}), \red{\forall k\in [K]}.
\end{array}
\end{align}

\begin{remark}[Conditional preservation of the Appendix-D attack-specific leakage bound]
\currentrev{For any feasible protected output, $\beta_t^{(k)}=h(\alpha_t^{(k)})=\|\alpha_t^{(k)}\|_2=\|W_t^{(k)}-\breve W_t^{(k)}\|_2$ holds by construction. For $I>0$, suppose the conditions of \bluerev{\pref{lem: bound_for_privacy_leakage}} hold with the same $c_a,c_b,D,c_2$, and $p$ uniformly over the feasible outputs under consideration. It then suffices to use a conservative lower-bound map satisfying}
{\color{red}
\begin{align*}
g^{(k)}(\chi_t^{(k)})\ge
\max\left\{\frac{2c_2c_b}{c_a}I^{p-1},
\frac{4D(1-\chi_t^{(k)})-c_2c_bI^{p-1}}{c_a}\right\}.
\end{align*}
}
Indeed, feasibility makes $\beta_t^{(k)}$ no smaller than the first term, so \bluerev{\pref{lem: bound_for_privacy_leakage}} applies; the second term then gives $\epsilon_p^{(k)}\le 1-(c_a\beta_t^{(k)}+c_2c_bI^{p-1})/(4D)\le\chi_t^{(k)}$. Thus, optimizing the distortion direction within this norm-feasible set does not invalidate the stated attack-specific bound. The notation $g^{(k)}$ suppresses its possible dependence on $t$: when a single map is reused over several rounds, the displayed constants must also be uniform over those rounds. This statement is conditional on the bi-Lipschitz and aggregate matching-residual assumptions; it does not assert a mechanism-independent privacy guarantee.
\end{remark}

Given a privacy budget, this formulation determines a lower bound on the distortion intensity needed to satisfy the privacy constraint. We therefore use \red{inequality constraints} rather than fixing the intensity at this lower bound, allowing the optimization to search over feasible distortion values up to the user-specified upper bound \(u_t^{(k)}\).

\textcolor{black}{The final optimization problem aims to derive the current-round distortion set $\{\alpha_t^{(k)}\}$ that balances utility and privacy. However, the problem is not guaranteed to be convex; it may be intractable and lack a closed-form solution. We therefore use adaptive SGD followed by an intensity-feasibility operation to approximately optimize it at each round, as detailed in \pref{sec: adaptive_learning_algorithm}.}

\section{Algorithm for Balancing Privacy and Utility} \label{sec: adaptive_learning_algorithm}




\currentrev{In this section, our framework approximately solves \pref{eq: optimization_problem_for_lemma} through constrained SGD followed by an intensity-feasibility operation. We first analyze the practical nonconvex $L_2$-annular projected update with a possibly stochastic or inexact implementation gradient, and then \bluerev{establish the exact projected-gradient $O(1/M)$ utility-loss gap as a convex-case corollary}.} The algorithm is shown in \pref{alg: adaLA}. We first present algorithmic procedures, and then provide formal guarantees on the per-round distortion subproblem and average stationarity.

\subsection{The Algorithm }
{\color{black}The main algorithm is presented in \pref{alg: adaLA}, with the distortion learning process shown in \pref{alg: LearnDistortionExtent}. The practical learner executes $M$ SGD steps on a sampled utility objective with an optional negative $L_2$-norm regularizer, applies an intensity-feasibility operation after each step, and returns the learned distortion. The operator $\operatorname{ScaleClip}_{h,[l,u]}$ radially rescales a nonzero input whose intensity is below $l$ or above $u$, and otherwise leaves it unchanged. This feasibility operation keeps the realized distortion intensity within the prescribed bounds. For the $L_2$-intensity PL branch, it is the Euclidean projection onto the nonconvex annulus analyzed in \pref{thm: distortion_stationarity_mt}; for the practical $L_1$ LS branch, it is a radial feasibility map rather than the Euclidean projection. For the monotone distribution-independent score in \pref{eq: measurement_for_privacy}, and under the uniform conditions stated in the remark following \pref{eq: optimization_problem_for_lemma}, feasibility preserves the configured attack-specific leakage constraint while the distortion direction is optimized. Hence, methods using the same score and budget enforce the same configured constraint; this comparison is not a mechanism-independent privacy guarantee. The key inputs include: 1) \textbf{\textit{privacy budget $\chi_t^{(k)}$}} for client $k$ at round $t$, 2) \textbf{\textit{the lower bound distortion extent $l_t^{(k)}$}} derived from the budget-to-distortion-extent mapping $g^{(k)}$, and 3) \textbf{\textit{the upper bound distortion extent}} $u_t^{(k)}$, which is heuristically set (e.g., two times the lower bound) to avoid excessive noise in practice. Two specific forms of $g^{(k)}$ are illustrated in \pref{exmp: distribution_free_prot_mechanism} and \pref{exmp: laplcian_prot_mechanism}.}


\begin{algorithm}[!htp]
    \caption{Privacy-preserving Federated Algorithm (PPFA)}
    \label{alg: adaLA}
    \begin{algorithmic}[1]
        \STATE \textbf{Parameters:} $T$: the number of training steps for the model parameter; \\
        \STATE \quad\enspace $W_0$: model parameter initialized by the server.
        \FOR{\red{$t=1,\ldots,T$}}
            \FOR{each client $k\in [K]$ \textbf{in parallel}}
              \STATE \textcolor{black}{Uniformly sample a mini-batch $\mathcal B_t^{(k)}$ from $\calD^{(k)}$}
              \STATE \textcolor{black}{$\breve W_t^{(k)}\leftarrow W_{t-1} -\eta\cdot\nabla \widehat{\calL}_{\mathcal B_t^{(k)}}^{(k)}(W_{t-1})$.} \quad \textcolor{black}{$\triangleright$ normal local model training}\\
              \STATE \textcolor{black}{$\alpha_t^{(k)}\leftarrow \text{LearnDistortion}(t,\breve W_t^{(k)},\mathcal B_t^{(k)})$.}  \quad \textcolor{black}{$\triangleright$ distortion learning}\\
              \STATE $\red{W_t^{(k)}}\leftarrow \red{\breve W_t^{(k)}} + \red{\alpha_t^{(k)}}$.\quad \textcolor{black}{$\triangleright$ apply distortion}\\
            \ENDFOR\\
            \textbf{Server executes:} \\
            \STATE $\red{W_t}\leftarrow \sum_{k = 1}^K p^{(k)}\red{W_t^{(k)}}$.
        \ENDFOR
    \end{algorithmic}
\end{algorithm}

%
%
%
    

\begin{exmp}[Distribution-independent Protection Mechanism]\label{exmp: distribution_free_prot_mechanism}
{\color{black}
For a distribution-independent protection mechanism, define $g^{(k)} (\chi^{(k)}\rightarrow \beta^{(k)})$. Using the attack-specific leakage score in \pref{eq: measurement_for_privacy}, we write
\begin{align}
   \epsilon_p^{(k)} (\beta^{(k)}) = a_1^{(k)} - a_2^{(k)}\beta^{(k)},
\end{align}
where $a_1^{(k)}=1-c_2c_bI^{p-1}/(4D)$, $a_2^{(k)}=c_a/(4D)$, and $\beta^{(k)}$ is the distortion intensity. On the nonvacuous parameter range of \bluerev{Lemma~\ref{lem: bound_for_privacy_leakage}}, these intercept and slope constants are positive. Because the lemma also requires $\beta^{(k)}\ge 2c_2c_bI^{p-1}/c_a$, a sufficient distortion lower bound for budget $\chi^{(k)}$ is
\begin{align}\label{eq: g_k_for_our_protection_mechanism}
g^{(k)}(\chi^{(k)})=\beta_{\min}^{(k)}
=\max\left\{
\frac{2c_2c_b}{c_a}I^{p-1},
\frac{a_1^{(k)}-\chi^{(k)}}{a_2^{(k)}}
\right\}.
\end{align}
Thus, even when the algebraic inverse is negative or smaller than the lemma threshold, Algorithm~\ref{alg: LearnDistortionExtent} uses a positive lower endpoint that discharges both requirements. The notation suppresses the round index: a round-specific map uses the constants conditioned on that round, while reusing one map across rounds requires those constants to hold uniformly over the relevant rounds. \bluerev{For every evaluated PL budget, the inverse term exceeds the lemma threshold, so the maximum above selects the inverse term.}
}
Furthermore, \red{$\alpha^{(k)}$ denotes the parameter-wise distortion, $\beta^{(k)}=h(\alpha^{(k)})$ denotes its scalar intensity, and $\epsilon_{u,t}^{(k)}(\alpha^{(k)}) = \calL^{(k)}(\breve W_t^{(k)} + \alpha^{(k)}) - \calL^{(k)}(\breve W_t^{(k)})$}, where $\mathcal{L}^{(k)}(W) = \frac{1}{|\calD^{(k)}|}\sum_{i\in\calD^{(k)}} \calL(W, d_i^{(k)})$, and $\calL(W, d_i^{(k)})$ represents the loss of predictions made by the model $W$ on $d_i^{(k)}$ ($i$-th data-label pair of client $k$).
\end{exmp}

\begin{exmp}[Distribution-dependent Protection Mechanism]\label{exmp: laplcian_prot_mechanism}
{
As in Example~\ref{exmp: distribution_free_prot_mechanism}, this generic calibration is stated for a fixed round, so the round subscript is suppressed while the client index remains the superscript $(k)$. Now we introduce $g^{(k)} (\chi^{(k)}\rightarrow \sigma^{(k)})$ for a distribution-dependent protection mechanism such as the Laplace mechanism. For the Laplace mechanism, the privacy leakage is measured using
\begin{align}
  \epsilon_{p}^{(k)}(\sigma^{(k)}) = \frac{S^{(k)}}{\sigma^{(k)}},
\end{align}
\red{where $S^{(k)}$ represents the $\ell_1$-sensitivity, and $\sigma^{(k)}$ is the scale of a zero-mean Laplace distribution $\operatorname{Lap}(0,\sigma^{(k)})$, with per-coordinate variance $2(\sigma^{(k)})^2$, and serves as the scalar distribution parameter}. Therefore, the distortion extent lower bound of a given budget $\chi^{(k)}$ is
\begin{align}\label{eq: g_k_for_dp_protection_mechanism}
   g^{(k)}(\chi^{(k)}) = \red{\sigma_{\min}^{(k)}} = \frac{S^{(k)}}{\chi^{(k)}}.
\end{align}
Furthermore, \red{$\sigma^{(k)}$ is the distribution scale and $\alpha_j^{(k)}\sim\operatorname{Lap}(0,\sigma^{(k)})$}\footnote{\red{Here, $\alpha^{(k)}$ is parameter-wise and $\sigma^{(k)}$ is the common Laplace scale of each parameter.}} and \red{$\epsilon_{u,t}^{(k)}(P_{\sigma^{(k)}}) = \E_{\alpha^{(k)}\sim P_{\sigma^{(k)}}}[\epsilon_{u,t}^{(k)}(\alpha^{(k)})]$}. \currentrev{This example specifies the idealized Laplace release and its scale-to-budget calibration. It does not by itself provide a formal differential-privacy guarantee for the adaptive realized-distortion learner used in Algorithm~\ref{alg: LearnDistortionExtent} and the experiments.}
}


\end{exmp}



\begin{algorithm}[!htp]
    \color{black}
    \caption{\textcolor{black}{LearnDistortion($t,\breve W_t^{(k)},\mathcal B_t^{(k)}$)}}
    \label{alg: LearnDistortionExtent}
   \begin{algorithmic}[1]
       \STATE \currentrev{\textbf{Parameters:} $M$: number of distortion-update steps; $\gamma_t$: learning rate;} 
       \STATE \currentrev{\quad$h$: distortion-intensity map; $\lambda_t^{(k)}\ge0$: negative-norm coefficient.}\\
    \STATE \textcolor{black}{\textbf{Initialization:}}
    \STATE \quad \textcolor{black}{$\beta_{t,\min}^{(k)}\leftarrow g^{(k)}(\chi_t^{(k)})$.} \quad \textcolor{black}{$\triangleright$ budget-induced lower bound}\\
    \STATE \quad \textcolor{black}{$l_t^{(k)}\leftarrow\beta_{t,\min}^{(k)}$, $u_t^{(k)}\leftarrow2\beta_{t,\min}^{(k)}$.} \quad \textcolor{black}{$\triangleright$ initialize intensity bounds}\\
    \STATE \quad \textcolor{black}{Draw the coordinates of $z^{(k)}$ from $\operatorname{Lap}(0,l_t^{(k)})$.}\\
    \IF{\textcolor{black}{$h(\alpha)=\|\alpha\|_2$ \textbf{(PL)}}}
       \STATE \quad \textcolor{black}{$\alpha_{t,1}^{(k)}\leftarrow l_t^{(k)}z^{(k)}/\|z^{(k)}\|_2$.} \quad \textcolor{black}{$\triangleright$ initialize on the inner sphere}
    \ELSE
       \STATE \quad \textcolor{black}{$\alpha_{t,1}^{(k)}\leftarrow z^{(k)}$; $l_t^{(k)}\leftarrow\|z^{(k)}\|_1$; $u_t^{(k)}\leftarrow2l_t^{(k)}$.} \quad \textcolor{black}{$\triangleright$ LS realized-norm bounds}
    \ENDIF
    \FOR{$m = 1, \cdots, M$}
       \STATE \currentrev{$G_{t,m}^{(k)}\leftarrow$ implementation gradient of}\\
       \STATE \quad \currentrev{$\widehat{\calL}_{\mathcal B_t^{(k)}}^{(k)}(\breve W_t^{(k)}+\alpha_{t,m}^{(k)})-\lambda_t^{(k)}\|\alpha_{t,m}^{(k)}\|_2$.} \quad \currentrev{$\triangleright$ possibly stochastic/inexact SGD direction}\\
       \STATE \currentrev{$\widetilde\alpha_{t,m+1}^{(k)}\leftarrow\alpha_{t,m}^{(k)}-\gamma_tG_{t,m}^{(k)}$.}\\
       \STATE \currentrev{$\alpha_{t,m+1}^{(k)}\leftarrow\operatorname{ScaleClip}_{h,[l_t^{(k)},u_t^{(k)}]}(\widetilde\alpha_{t,m+1}^{(k)})$.} \quad \currentrev{$\triangleright$ enforce intensity bounds}
    \ENDFOR
    \STATE \textcolor{black}{\textbf{return} $\alpha_t^{(k)}\leftarrow\alpha_{t,M+1}^{(k)}$}
   \end{algorithmic}
\end{algorithm}


\subsection{Main Results}

{\color{black}
\currentrev{For a fixed round $t$, client $k$, and realized mini-batch $\mathcal B_t^{(k)}$, define the full-client utility objective, the sampled utility objective, and the practical regularized sampled objective by}
{\color{red}
\begin{align}
f_t^{(k)}(\alpha)
&\triangleq\epsilon_{u,t}^{(k)}(\alpha)
=\calL^{(k)}(\breve W_t^{(k)}+\alpha)-\calL^{(k)}(\breve W_t^{(k)}),\\
\widehat f_{t,\mathcal B}^{(k)}(\alpha)
&\triangleq\widehat{\calL}_{\mathcal B_t^{(k)}}^{(k)}
(\breve W_t^{(k)}+\alpha)
-\widehat{\calL}_{\mathcal B_t^{(k)}}^{(k)}(\breve W_t^{(k)}),\\
\phi_{t,\mathcal B}^{(k)}(\alpha)
&\triangleq\widehat f_{t,\mathcal B}^{(k)}(\alpha)
-\lambda_t^{(k)}\|\alpha\|_2,
\qquad \lambda_t^{(k)}\ge0,
\end{align}
}
where $\widehat{\calL}_{\mathcal B_t^{(k)}}^{(k)}$ is the empirical loss on the realized mini-batch. The final term is the small negative-norm regularizer used by the practical learner; setting $\lambda_t^{(k)}=0$ recovers pure sampled utility minimization. For the $L_2$-intensity PL realization, the feasible set is
{\color{red}
\begin{align}
\mathcal A_t^{(k)}
\triangleq
\left\{\alpha:
l_t^{(k)}\le\|\alpha\|_2\le u_t^{(k)}
\right\}.
\label{eq: l2_annular_feasible_set}
\end{align}
}
\currentrev{When $l_t^{(k)}>0$, this closed annulus is nonconvex. The first result below concerns the inner distortion-update trajectory for this fixed round, client, and sampled subproblem, rather than global optimality of the complete federated trajectory.}

For the global convergence result, let $F(W)\triangleq\sum_{k=1}^Kp^{(k)}\calL^{(k)}(W)$, let $g_t$ denote the aggregated stochastic gradient at $W_{t-1}$, and recall $\Delta_t=\sum_{k=1}^Kp^{(k)}\alpha_t^{(k)}$. The server update can then be written as
\begin{align}\label{eq: perturbed_fedsgd_update}
    W_t=W_{t-1}-\eta g_t+\Delta_t.
\end{align}
}

{\color{black}
\begin{assumption}[Nonconvex distortion subproblem and implementation gradient]
\label{assump: nonconvex_distortion_subproblem}
For the round, client, and realized mini-batch under consideration, assume $0<l_t^{(k)}<u_t^{(k)}<\infty$, and abbreviate $\mathcal A=\mathcal A_t^{(k)}$ and $\phi=\phi_{t,\mathcal B}^{(k)}$. Let $\phi_{\inf}\triangleq\inf_{\alpha\in\mathcal A}\phi(\alpha)>-\infty$. There exists $L_\alpha>0$ such that, for every $x,y\in\mathcal A$,
\begin{align}
\phi(y)&\le\phi(x)+\langle\nabla\phi(x),y-x\rangle
+\frac{L_\alpha}{2}\|y-x\|_2^2,
\label{eq: restricted_smooth_descent}\\
\|\nabla\phi(y)-\nabla\phi(x)\|_2
&\le L_\alpha\|y-x\|_2.
\label{eq: restricted_gradient_lipschitz}
\end{align}
{\color{red}
Let $\mathcal H_m$ contain the realized mini-batch, the initialization, and the inner-loop history through $\alpha_m$. If $G_m$ denotes the possibly stochastic or inexact gradient produced at inner step $m$, assume
\begin{align}
\E\!\left[
\|G_m-\nabla\phi(\alpha_m)\|_2^2
\mid\mathcal H_m
\right]
\le\nu_\alpha^2.
\label{eq: inner_gradient_mse}
\end{align}
No conditional-unbiasedness assumption is imposed on $G_m$.
}
\end{assumption}
}

\currentrev{For $h(\alpha)=\|\alpha\|_2$, the norm-scaling and norm-clipping operation used by PL-Learn is the Euclidean projection onto \pref{eq: l2_annular_feasible_set}, up to its numerical stabilizer.} At a zero trial point, the theoretical operator selects any fixed point on the inner sphere. The theorem below also applies to another intensity map whenever its implemented operation is an exact Euclidean projection. \currentrev{In particular, it does not claim to cover the practical $L_1$ radial-scaling operation used by LS-Learn.}

\begin{assumption}[Global objective]\label{assump: global_objective}
\yellow{The global objective $F$ is lower bounded by $F_*$ and has an $L_F$-Lipschitz gradient.}
\end{assumption}

\begin{assumption}[Stochastic gradient and distortion]\label{assump: stochastic_gradient_distortion}
\currentrev{Let $W_0$ be deterministic, and let $\mathcal F_{t-1}$ contain $W_0$ and all randomness revealed through round $t-1$, so that $W_{t-1}$ is $\mathcal F_{t-1}$-measurable. The aggregated stochastic gradient satisfies}
{\color{red}
\begin{align}
    \E[g_t\mid\mathcal F_{t-1}]&=\nabla F(W_{t-1}),\\
    \E[\|g_t-\nabla F(W_{t-1})\|^2\mid\mathcal F_{t-1}]&\le\sigma_g^2.
\end{align}
}
The variables $g_t$, $\Delta_t$, and $W_t$ are adapted to the round-wise filtration and have the finite moments used below. \currentrev{The aggregate distortion may be adaptive or biased, but $\E[\|\Delta_t\|^2]<\infty$.}
\end{assumption}

\currentrev{The following theorem directly analyzes the nonconvex projected-SGD update used by the practical $L_2$-annular learner.}
{\color{black}
\begin{thm}\label{thm: distortion_stationarity_mt}
Let $M\ge1$, fix $t$, $k$, and $\mathcal B_t^{(k)}$, and let \pref{assump: nonconvex_distortion_subproblem} hold. Suppose $\alpha_1\in\mathcal A$ and, for $m=1,\ldots,M$, the implementation uses a constant step size $0<\gamma_t<1/L_\alpha$ and any fixed measurable Euclidean-projection selection
\begin{align}
\alpha_{m+1}
\in\operatorname{Proj}_{\mathcal A}
\left(\alpha_m-\gamma_tG_m\right).
\label{eq: nonconvex_projected_sgd_update}
\end{align}
Let $N_{\mathcal A}(\alpha)$ denote the limiting normal cone of $\mathcal A$ at $\alpha$, and define
\begin{align}
R_m&\triangleq\frac{\alpha_m-\alpha_{m+1}}{\gamma_t},\\
\mathfrak s(\alpha)&\triangleq
\operatorname{dist}\!\left(
0,\nabla\phi(\alpha)+N_{\mathcal A}(\alpha)
\right),\\
\Delta_\phi&\triangleq
\E[\phi(\alpha_1)]-\phi_{\inf}.
\end{align}
Then the selected projected-step residual satisfies
\begin{align}
\frac1M\sum_{m=1}^M\E[\|R_m\|_2^2]
\le
\underbrace{
\frac{4\Delta_\phi}
{\gamma_t(1-L_\alpha\gamma_t)M}
+\frac{4\nu_\alpha^2}
{(1-L_\alpha\gamma_t)^2}
}_{\displaystyle \mathcal E_M},
\label{eq: nonconvex_step_stationarity}
\end{align}
and its limiting-normal-cone stationarity satisfies
\begin{align}
\frac1M\sum_{m=1}^M
\E[\mathfrak s(\alpha_{m+1})^2]
\le
2(1+L_\alpha\gamma_t)^2\mathcal E_M
+2\nu_\alpha^2.
\label{eq: nonconvex_normal_stationarity}
\end{align}
If $G_m=\nabla\phi(\alpha_m)$, the constants improve to
\begin{align}
\frac1M\sum_{m=1}^M\E[\|R_m\|_2^2]
&\le
\frac{2\Delta_\phi}
{\gamma_t(1-L_\alpha\gamma_t)M},
\label{eq: exact_nonconvex_step_stationarity}\\
\frac1M\sum_{m=1}^M
\E[\mathfrak s(\alpha_{m+1})^2]
&\le
\frac{2(1+L_\alpha\gamma_t)^2\Delta_\phi}
{\gamma_t(1-L_\alpha\gamma_t)M}.
\label{eq: exact_nonconvex_normal_stationarity}
\end{align}
\end{thm}
}

{\color{black}
\currentrev{Thus, the practical nonconvex inner trajectory has an $O(1/M)$ transient term and an $O(\nu_\alpha^2)$ stationary-neighborhood term. The result controls the trajectory average, the best inner iterate, or an iterate selected uniformly from the trajectory; it does not assert a last-iterate rate or global optimality for the nonconvex annular problem.}

\begin{cor}[Convex exact-gradient case]\label{cor: convex_suboptimality_mt}
Suppose instead that $\mathcal C_t^{(k)}$ is a nonempty closed convex feasible set with finite diameter $D_t^{(k)}$, $\lambda_t^{(k)}=0$, $f_t^{(k)}$ is convex with an $L_\alpha$-Lipschitz gradient, and exact projected-gradient steps with $0<\gamma_t\le1/L_\alpha$ are used. For any
\begin{align}
\alpha_t^{(k)*}\in
\arg\min_{\alpha\in\mathcal C_t^{(k)}}f_t^{(k)}(\alpha),
\end{align}
the final inner iterate obeys
\begin{align}\label{eq: per_round_suboptimality}
0&\le
\epsilon_{u,t}^{(k)}(\alpha_{t,M+1}^{(k)})
-\epsilon_{u,t}^{(k)}(\alpha_t^{(k)*})\\
&\le
\frac{\|\alpha_{t,1}^{(k)}-\alpha_t^{(k)*}\|_2^2}
{2\gamma_tM}
\le\frac{(D_t^{(k)})^2}{2\gamma_tM}.
\end{align}
\bluerev{Consequently, the global utility-loss gap decreases as $O(1/M)$ in this convex exact-gradient regime. This corollary serves as a benchmark for the convex case; it is not invoked for the positive-lower-bound annulus or the regularized practical objective.}
\end{cor}
}

\bluerev{Theorem~\ref{thm: distortion_stationarity_mt} and Corollary~\ref{cor: convex_suboptimality_mt} concern the inner distortion subproblem.} \currentrev{The following theorem separately characterizes the average stationarity of the global federated-model trajectory.} The average squared gradient norm \cite{lian2015asynchronous,zeng2018nonconvex,bottou2018optimization,wang2018cooperative}, commonly considered as a convergence indicator, is utilized.
\begin{thm}\label{thm: converge_rate_mt}
\currentrev{Let \pref{assump: global_objective} and \pref{assump: stochastic_gradient_distortion} hold, and choose $0<\eta\le1/(4L_F)$. The average squared gradient norm of our proposed algorithm is bounded by}
{\color{red}
\begin{align}
\frac{1}{T}\sum_{t=1}^T\E[\|\nabla F(W_{t-1})\|^2]
&\le\frac{2(F(W_0)-\E[F(W_T)])}{\eta T}\nonumber\\
&\quad+\frac{2}{T}\sum_{t=1}^T
\left[
L_F\eta\sigma_g^2
+\left(\frac{1}{\eta^2}+\frac{L_F}{\eta}\right)
\E[\|\Delta_t\|^2]
\right].\label{eq: converge_rate_mt}
\end{align}
}
Since $F$ is lower bounded by $F_*$, the transient term in \pref{eq: converge_rate_mt} can be upper bounded by $2(F(W_0)-F_*)/(\eta T)$. In particular, if $\E[\|\Delta_t\|^2]\le\rho^2$ for all $t$, then
\begin{align}
\limsup_{T\to\infty}\frac{1}{T}\sum_{t=1}^T\E[\|\nabla F(W_{t-1})\|^2]
\le2L_F\eta\sigma_g^2
+2\left(\frac{1}{\eta^2}+\frac{L_F}{\eta}\right)\rho^2.
\end{align}
\end{thm}

This theorem bounds the average squared gradient norm by the transient optimization term, stochastic-gradient variance, and the expected squared norm of the aggregate distortion. \currentrev{With fixed nonzero distortion, the bound describes a stationary neighborhood; it does not imply convergence of the individual iterates or gradients.}

\currentrev{Lowy and Razaviyayn~\cite{lowy2023private} provide excess-risk guarantees for heterogeneous federated learning under inter-silo record-level differential privacy without a trusted server, assuming convex and smooth losses. These guarantees provide a privacy--utility benchmark for DP-calibrated mechanisms, whereas \pref{thm: converge_rate_mt} characterizes the optimization effect of aggregate distortion through the average squared gradient norm; thus, the two results are complementary rather than directly interchangeable.}






\setlength{\belowcaptionskip}{0pt}
\setlength{\abovecaptionskip}{0pt}

\newcommand{\bluefiglabel}[2]{%
  \begingroup\setlength{\fboxsep}{0.5pt}%
  \colorbox{white}{\makebox[#1][l]{\textcolor{black}{#2}}}%
  \endgroup}

\section{Experiments}

\begin{figure}[t]
    \centering
    \includegraphics[width=0.49\textwidth]{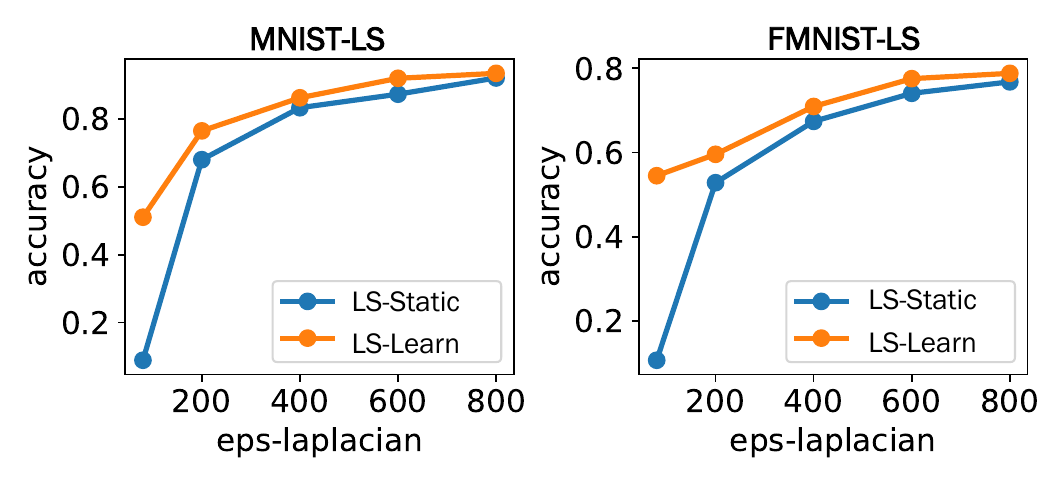}
    \includegraphics[width=0.49\textwidth]{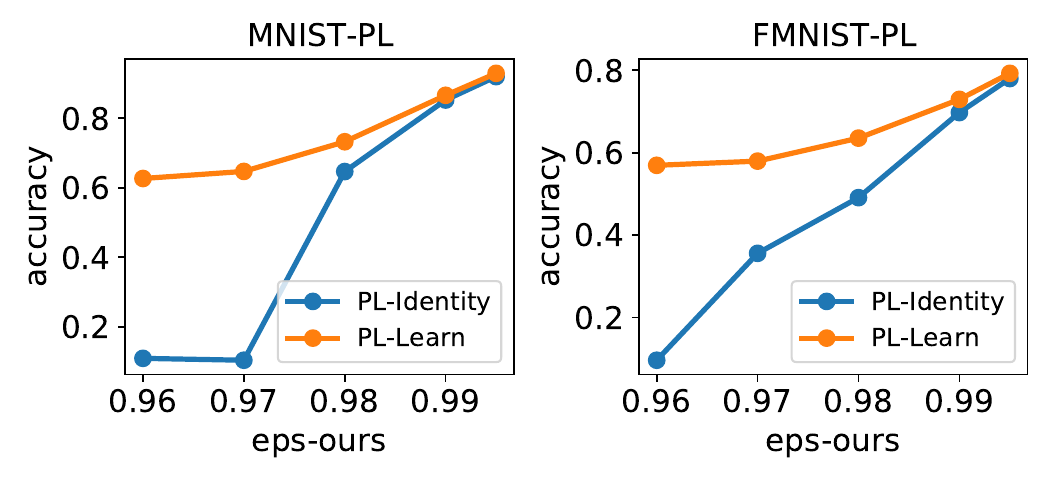}
    \caption{The privacy-utility trade-off curve of different defense mechanisms. } 
    \label{fig:eps-acc}
\end{figure}

\begin{table}[t]
\centering
\caption{The detailed privacy-utility performance under the \textcolor{black}{Laplace-scale calibration setting}.} 
\begin{tabular}{@{}c|r|ccccc@{}}
\toprule
Dataset                 & \multicolumn{1}{c|}{budget($\chi$)} & 80               & 200             & 400             & 600             & 800             \\ \midrule
\multirow{3}{*}{MNIST}  & Identical                    & 8.92\%           & 67.98\%         & 83.28\%         & 87.26\%         & 92.04\%         \\
                        & Learn                        & 51.04\%          & 76.46\%         & 86.20\%         & 91.94\%         & 93.40\%         \\ \cmidrule(l){2-7} 
                        & diff                         & \textbf{42.12\%} & \textbf{8.48\%} & \textbf{2.92\%} & \textbf{4.68\%} & \textbf{1.36\%} \\ \midrule
\multirow{3}{*}{FMNIST} & Identical                    & 10.68\%          & 52.82\%         & 67.38\%         & 74.04\%         & 76.76\%         \\
                        & Learn                        & 54.48\%          & 59.56\%         & 70.92\%         & 77.52\%         & 78.78\%         \\ \cmidrule(l){2-7} 
                        & diff                         & \textbf{43.80\%} & \textbf{6.74\%} & \textbf{3.54\%} & \textbf{3.48\%} & \textbf{2.02\%} \\ \bottomrule
\end{tabular}
\label{tab:dp-eps-acc}
\end{table}

\begin{table}[t]
\centering
\caption{Utility-Privacy Trade-off under our own privacy framework. Our method consistently outperforms the baseline.}
\begin{tabular}{@{}c|r|ccccc@{}}
\toprule
Dataset                 & \multicolumn{1}{c|}{budget($\chi$)} & 0.96    & 0.97    & 0.98    & 0.99    & 0.995   \\ \midrule
\multirow{3}{*}{MNIST}  & Identical                    & 10.92\% & 10.38\% & 64.68\% & 85.26\% & 92.00\% \\
                        & Learn                        & 62.68\% & 64.72\% & 73.26\% & 86.62\% & 92.94\% \\ \cmidrule(l){2-7} 
 & diff & \textbf{51.76\%} & \textbf{54.34\%} & \textbf{8.58\%}  & \textbf{1.36\%} & \textbf{0.94\%} \\ \midrule
\multirow{3}{*}{FMNIST} & Identical                    & 9.64\%  & 35.58\% & 49.10\% & 69.74\% & 78.00\% \\
                        & Learn                        & 56.94\% & 57.96\% & 63.54\% & 72.94\% & 79.26\% \\ \cmidrule(l){2-7} 
 & diff & \textbf{47.30\%} & \textbf{22.38\%} & \textbf{14.44\%} & \textbf{3.20\%} & \textbf{1.26\%} \\ \bottomrule
\end{tabular}
\label{tab:pl-eps-acc}
\end{table}

\subsection{Experiment Setup}
We conducted experiments on MNIST and Fashion-MNIST to evaluate the effectiveness of our general distortion learning algorithm. The detailed federated training and attack settings are described below. The experiments are designed to answer the following research questions. \textbf{\underline{Q1}}: Does the proposed method achieve a better utility-privacy trade-off than static methods? \textbf{\underline{Q2}}: Does the proposed method perform consistently under different privacy measures? 

\paragraph{\textbf{FL Setting}} To avoid the introduction of overly complex factors and evaluate the effectiveness of our framework itself, we employ a simple and straightforward FL setting for experiments. \textcolor{black}{We use FedSGD with full client selection, one local-batch update step per round, and a local batch size of 4.} As a mild choice, the number of clients is 4 in all experiments. For each client, there are 1K, 1.2K and 1.2K samples for training, validation, and testing, respectively, and the label distributions of all clients are identical. We follow \cite{zhu2020deep,geiping2020inverting} to use a simplified LeNet Model (denoted as ``LeNetZhu'') as the backbone network. It has 3 convolution layers and 1 fully connected layer. 

\paragraph{\textbf{Attack Setting}} We select InvGrad \cite{geiping2020inverting} as the attacking method. As a general attack assumption for FL, the attacker is assumed to have knowledge of the weight updates \cite{geiping2020inverting}. We also assume the attacker has knowledge of the true label to simplify the attack. For each target batch, 1600 attack iterations are conducted with the Adam optimizer, the attack learning rate is set to 1, and the total variation coefficient is 1e-5. \textcolor{black}{Due to the limitations of gradient-inversion methods in large-batch attacks, each attacked update is generated from the first local mini-batch of four examples at a converged round, and InvGrad reconstructs that same four-example target batch.}

\paragraph{\textbf{Defense Setting}}
We set up defense methods for both distribution-independent (e.g. Eq(\ref{eq: g_k_for_our_protection_mechanism})) and distribution-dependent (e.g. Eq(\ref{eq: g_k_for_dp_protection_mechanism})) mechanisms to validate the effectiveness of our algorithm under distinct privacy leakage measurements. They are compared under identical FL and attack settings. Under each privacy framework, we set up a degraded version of our algorithm using a traditional static distortion strategy as the baseline. The four defense methods are listed below:
\begin{itemize}[leftmargin=*]
    \item \textbf{{PL-Learn}}: \textcolor{black}{The complete version of \pref{alg: LearnDistortionExtent} with our \underline{\textbf{P}}rivacy \underline{\textbf{L}}eakage definition. For each model, the distortion is dynamically adapted in the range $[l_t^{(k)},u_t^{(k)}]$ by the nonconvex $L_2$-annular projected-SGD learner.}
    \item \textbf{{PL-Identical}}: The degraded version of PL-Learn, a random noise with lower-bound noise extent (decided by $l_t^{(k)}$ in \pref{alg: LearnDistortionExtent}) is uniformly applied to each model parameter, without optimization. Here, ``Identical'' refers to the use of a fixed global $L_2$ distortion intensity at the lower bound, rather than identical values in individual coordinates or a deterministic distortion direction.
    \item \textbf{\textcolor{black}{LS-Learn}}: The complete version of \pref{alg: LearnDistortionExtent} under the \textcolor{black}{Laplace-scale calibration setting}. The algorithm optimizes the \textcolor{black}{realized distortion initialized from a random vector} to achieve a better privacy-utility trade-off.
    \item \textbf{\textcolor{black}{LS-Static}}: It is the degraded version of \textcolor{black}{LS-Learn}, similar to PL-Identical.
\end{itemize}

\noindent \textbf{Budget Setting}: For each type of protection mechanism, we adopt a consistent range of budget values to enable a comprehensive evaluation of the utility–privacy trade-off. This range spans from configurations that cause noticeable utility degradation to those with only minimal impact. However, choosing an appropriate budget is non-trivial in practice. It depends on multiple factors such as the dataset, model architecture, and initialization strategy. Excessive noise may prevent the model from converging, while insufficient noise offers little insight into a mechanism’s effectiveness. To identify suitable budgets, we perform empirical utility testing for each dataset to assess the model’s tolerance to different noise magnitudes. \textcolor{black}{\bluerev{The PL calibration follows Example~\ref{exmp: distribution_free_prot_mechanism}, whereas the LS calibration follows Example~\ref{exmp: laplcian_prot_mechanism}. For the PL parameters $(D,c_a,c_{\mathrm{res}},p,I)=(56,0.56,0.01,1/2,1600)$, where $c_{\mathrm{res}}:=c_2c_b/c_a$, the lemma-imposed threshold is $5\times10^{-4}$ and the inverse term is at least $1.99975$ over the evaluated range $\chi\in[0.96,0.995]$. Therefore, the maximum in Example~\ref{exmp: distribution_free_prot_mechanism} selects the inverse term for every evaluated PL setting; this arithmetic check does not verify Assumption~\ref{assump: matching_residual}.}} Through this procedure, we find that model training fails when the budget drops below 0.96 for PL-Identical and 80 for \textcolor{black}{LS-Static}. These thresholds are therefore set as the minimum acceptable budgets. From there, we incrementally increase the values until the added noise has negligible impact on utility. Finally, although specific budget values affect model performance significantly, our evaluation criterion is orthogonal to this choice: regardless of the selected budget, the protection mechanism that achieves a better utility–privacy trade-off is considered more effective.

\paragraph{\textbf{Evaluation Setting}}
The test accuracy of the global model is used as the utility metric. For each privacy framework, we set progressive budget values and evaluate utility of defense methods. Thus, the algorithm is considered to maintain a better trade-off when it achieves better utility performance \textcolor{black}{at fixed budgets}. To empirically evaluate the privacy protection effect of defense mechanisms, we also use MSE and SSIM to assess the quality of \textcolor{black}{InvGrad-reconstructed images}. Additionally, in this work, we utilize Calibrated Averaged Performance (CAP) \cite{fan2020rethinking} as a measure to quantify the balance between the accuracy of the main task and the error in data recovery. Let $Acc(\cdot)$ represent the accuracy of the main task, and $Rerr(\cdot)$ denote the recovery error between the original data $x$ and the estimated data $\hat{x_s}$ obtained through the attack $a$. CAP is defined for a given defense mechanism $g_s \in G$ (where $s$ represents the controlled parameter of $g$, such as the noise level or budget level) and an attacking mechanism $a \in \mathcal{A}$

$$
\operatorname{CAP}\left(g_s, a\right)=\frac{1}{m} \sum_{s=s_1}^{s_m} \operatorname{Acc}\left(g_s, x\right) * \operatorname{Rerr}\left(x, \hat{x}_s\right)
$$

\begin{table}[t]
\begin{subtable}[h]{0.48\textwidth}
\centering
\begin{tabular}{@{}c|c|cc@{}}
\toprule
\multicolumn{1}{l|}{Dataset} & Metric & \textcolor{black}{LS}      & PL      \\ \midrule
\multirow{2}{*}{MNIST}       & mse    & 0.002   & -0.078  \\
                             & ssim   & 0.00\%  & -0.31\% \\ \midrule
\multirow{2}{*}{FMNIST}      & mse    & -0.044  & 0.015   \\
                             & ssim   & -0.54\% & -0.17\% \\ \bottomrule
\end{tabular}
\caption{The average gap in defense effect (measured by MSE and SSIM) between our method and the baseline. Differences close to zero indicate similar empirical protection \textcolor{black}{across the tested budgets}. Detailed results are shown in Table \ref{tab:dp-mse} and Table \ref{tab:pl-mse}.}
\label{tab:emprical-privacy}
\end{subtable}
\hfill
\begin{subtable}[t]{0.48\textwidth}
\centering
\begin{tabular}{@{}c|c|cc@{}}
\toprule
Dataset                 & Method    & \textcolor{black}{LS}              & PL              \\ \midrule
\multirow{3}{*}{MNIST}  & Identical & 1.62            & 1.30            \\
                        & Learn     & 2.02            & 2.09            \\ \cmidrule(l){2-4} 
                        & up ratio  & \textbf{24.9\%} & \textbf{61.3\%} \\ \midrule
\multirow{3}{*}{FMNIST} & Identical & 1.08            & 0.91            \\
                        & Learn     & 1.31            & 1.27            \\ \cmidrule(l){2-4} 
                        & up ratio  & \textbf{21.2\%} & \textbf{39.1\%} \\ \bottomrule
\end{tabular}
\caption{\textcolor{black}{The Calibrated Averaged Performance (CAP) of various protection mechanisms against the InvGrad attack.}}
\label{tab:cap}
\end{subtable}
\caption{Summary comparison of the empirical utility and privacy performance of our method.}
\end{table}

\begin{figure*}[t]
    \centering
    \includegraphics[width=\textwidth]{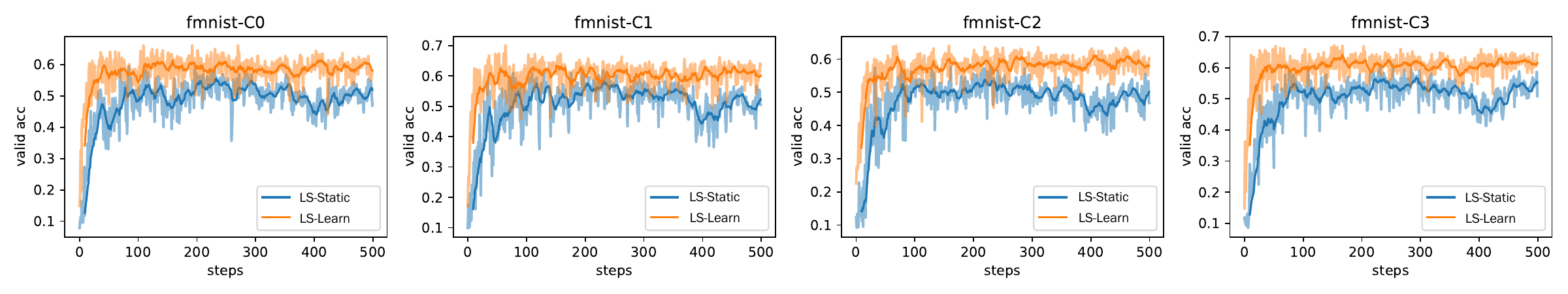}
    \includegraphics[width=\textwidth]{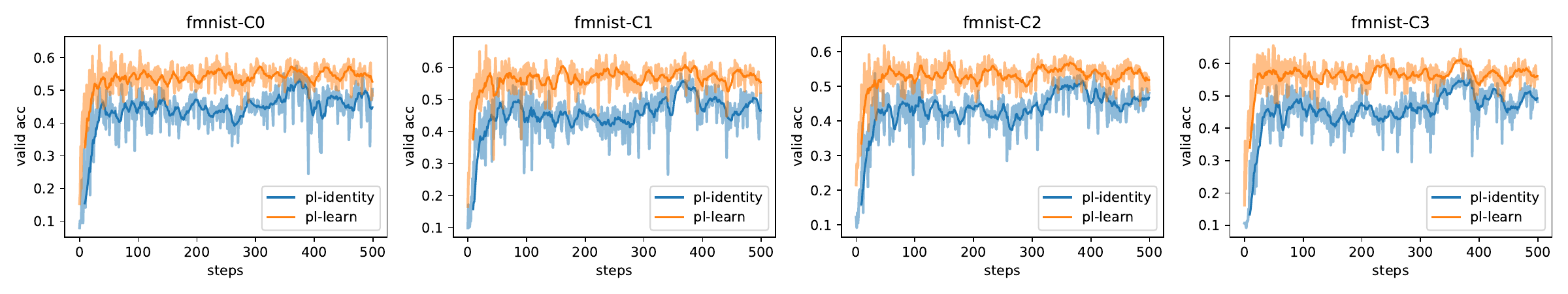}
\caption{The client training process of our methods (the $1_{st}$ row shows \textcolor{black}{$\chi_{\mathrm{LS}}=200$}, and the $2_{nd}$ row shows \textcolor{black}{$\chi_{\mathrm{PL}}=0.975$}) on Fashion-MNIST. All clients achieve better convergence.}
    \label{fig:clients}
\end{figure*}

\begin{figure}[t]
    \centering
    \includegraphics[width=0.49\textwidth]{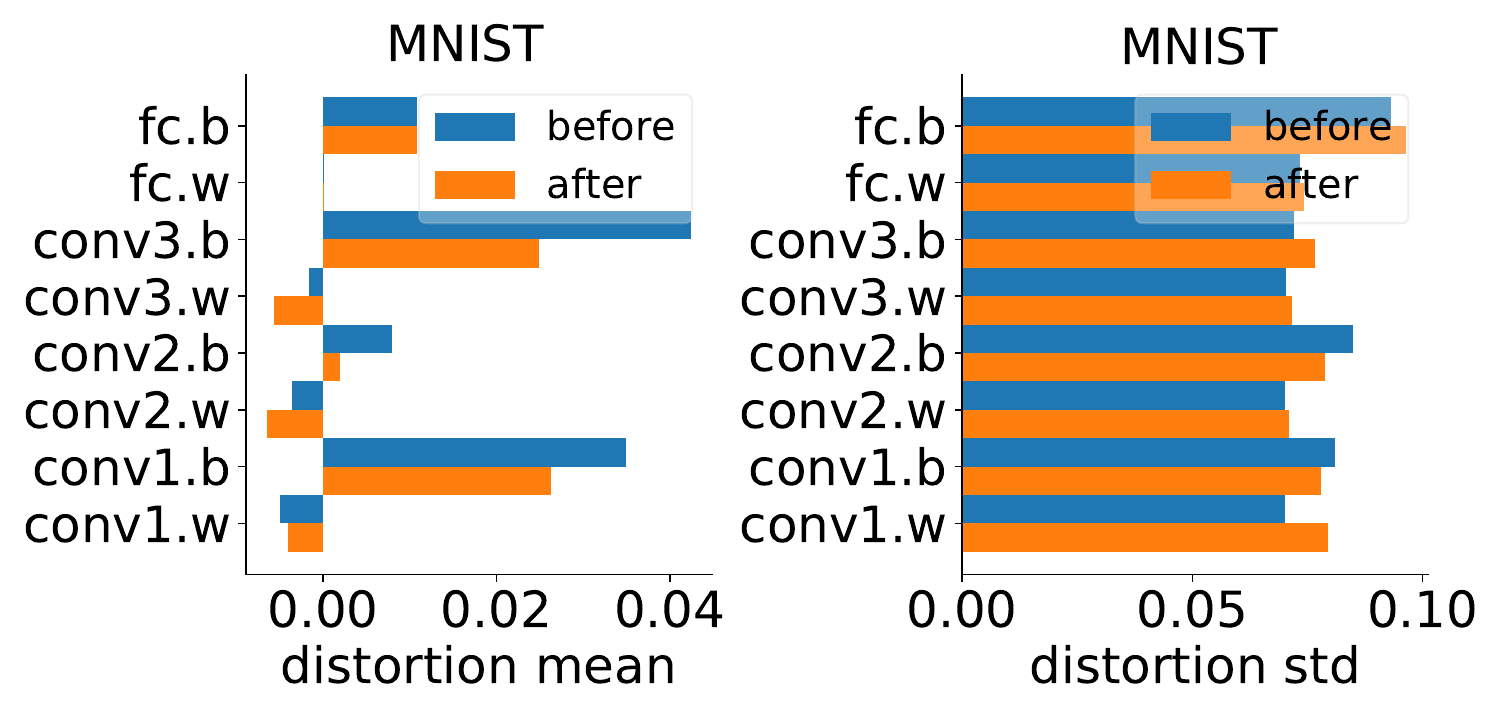}
    \includegraphics[width=0.49\textwidth]{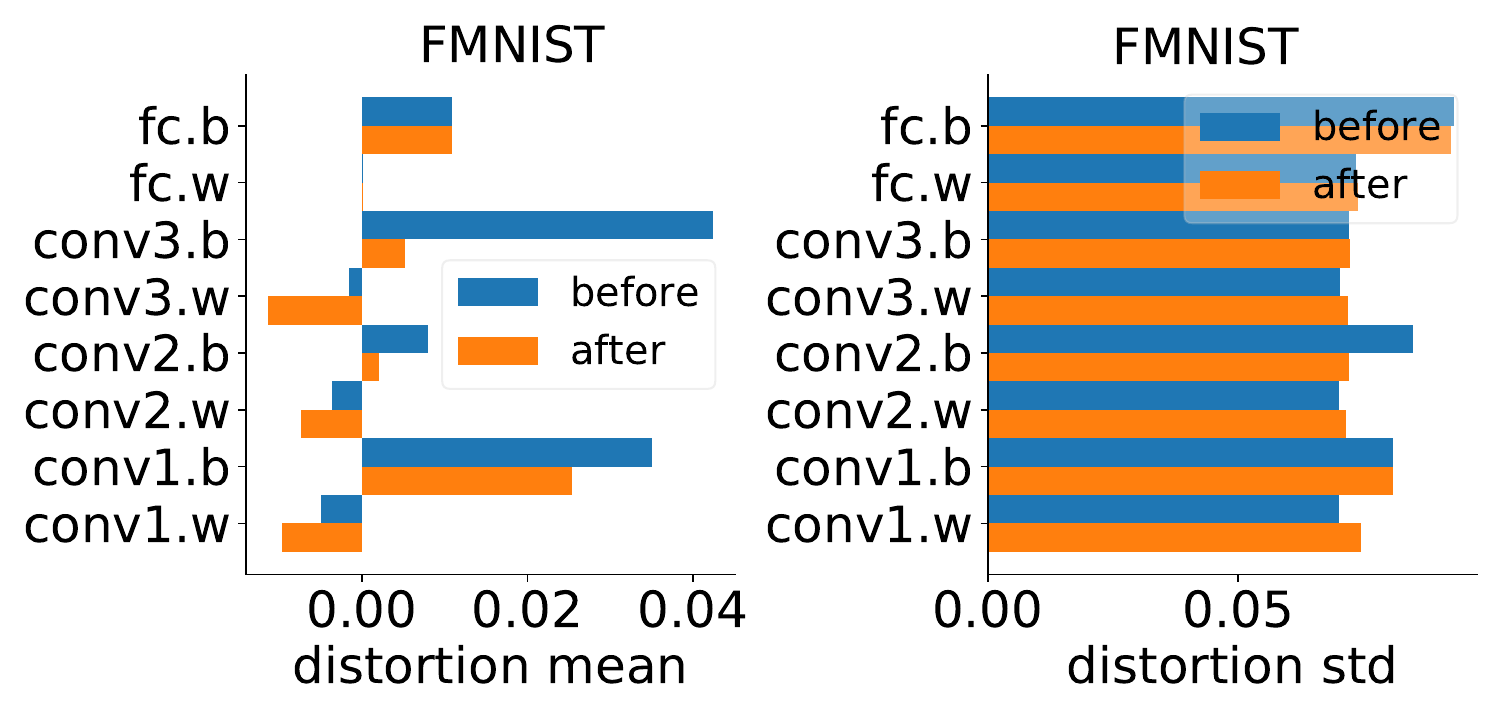}
    \caption{The impact of learning process on distortion values. The algorithm selectively amplifies(conv2.w, conv3.b), reduces(conv1.w, conv1.b, conv2.b, conv3.b) and maintains (fc.b, fc.w) the distortion extent of different layers to boost utility (in the $1$st sub-figure), \textcolor{black}{while the layer-wise standard deviations change only slightly across the tested settings (in the $2$nd sub-figure)}.}
    \label{fig:delta-stat}
\end{figure}

\begin{figure}[t]
    \centering
    \includegraphics[width=0.49\textwidth]{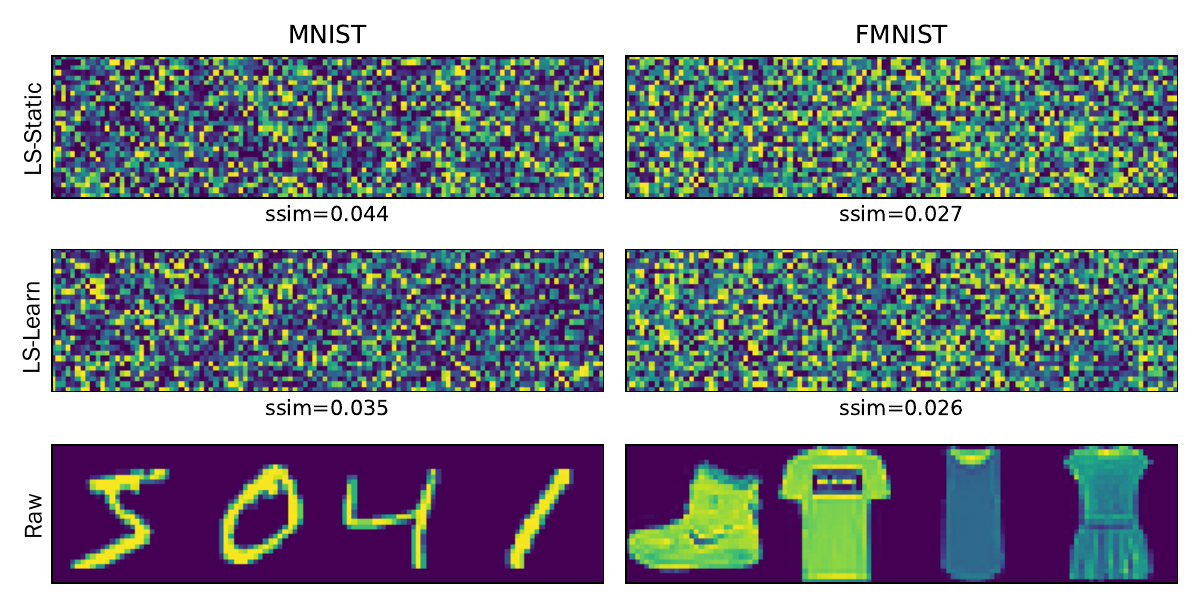}
    \includegraphics[width=0.49\textwidth]{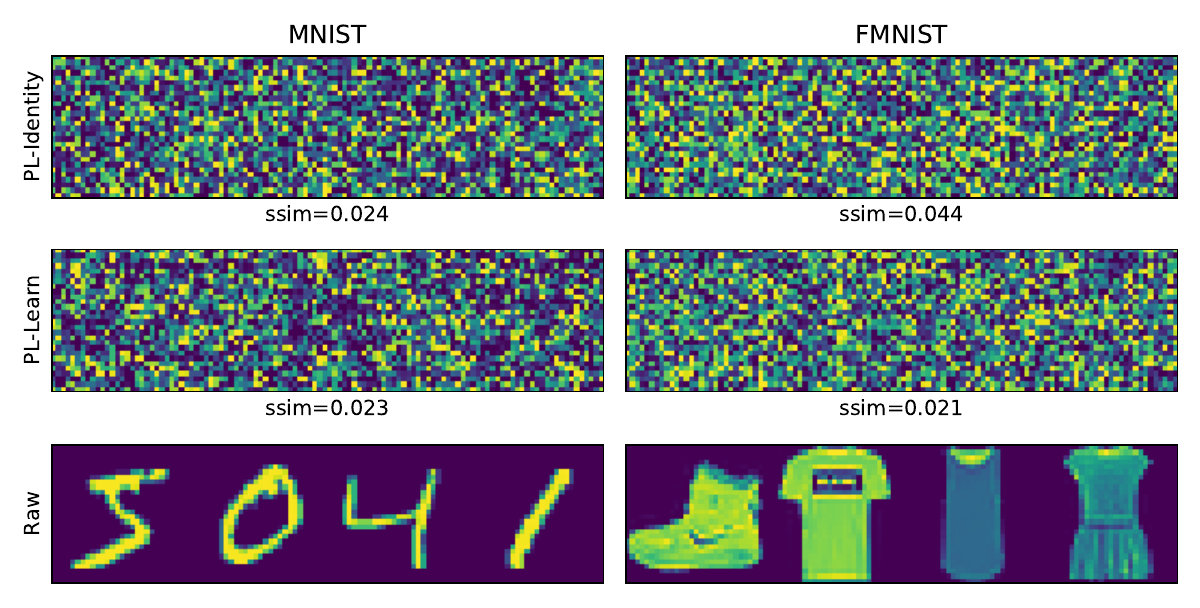}
\caption{\textcolor{black}{InvGrad reconstruction results under representative Laplace-scale and distribution-independent calibration settings (with $\chi_{\mathrm{LS}}=200$ and $\chi_{\mathrm{PL}}=0.98$, respectively). Our method and the baseline exhibit comparable empirical protection quantitatively and visually.}}
    \label{fig:dlg-rec-img}
\end{figure}

\paragraph{\textbf{Implementation Details}} \textcolor{black}{For the Laplace-scale calibration experiment, we use threshold $C=500$ for gradient clipping to bound the sensitivity used in the scale mapping (see, e.g., \cite{abadi2016deep, zhu2021fine}).} \textcolor{black}{\currentrev{For distortion learning, we use plain SGD with learning rate $\gamma_t=0.1$ for $M=10$ inner steps and reuse the realized local mini-batch throughout those steps.} The implementation initializes the distortion coordinate-wise from $\operatorname{Lap}(0,l_t^{(k)})$. PL-Learn rescales this sample to the $L_2$ lower boundary, whereas LS-Learn retains the sampled distortion and sets its realized $L_1$ interval to $[\|\alpha_{t,1}^{(k)}\|_1,2\|\alpha_{t,1}^{(k)}\|_1]$. The learner also optimizes the sampled utility objective with a negative $L_2$-norm coefficient $\lambda_t^{(k)}=10^{-5}$. \currentrev{Although the initialization is Laplace, the subsequently learned parameter-wise distortion is not constrained to remain i.i.d. Laplace distributed. Thus, the LS experiments evaluate Laplace-scale-calibrated realized distortion rather than a formal differential-privacy release.}}
\textcolor{black}{\currentrev{For PL-Learn, norm scaling/clipping is, up to the $10^{-9}$ numerical stabilizer, the Euclidean projection onto the nonconvex $L_2$ annulus in \pref{eq: l2_annular_feasible_set}; this is the update structure analyzed in \pref{thm: distortion_stationarity_mt}, whose mean-square oracle-error term permits a stochastic or inexact implementation direction. For LS-Learn, the implementation uses $L_1$ radial scaling as a practical feasibility map, so the exact-Euclidean-projection theorem is not invoked for that branch.} The experiments instantiate the practical update structure but are not presented as a direct empirical verification of the stationarity bound. Moreover, the $O(1/M)$ global utility-loss gap in \pref{cor: convex_suboptimality_mt} is only a convex exact-gradient benchmark and is not claimed for the positive-lower-bound annular experiments.} For the common training settings, we use the same random seed $seed=1$ and the same \textcolor{black}{InvGrad initialization} in all experiments except the average-squared-gradient-norm experiment in Figure \ref{fig:average_squared_gradient_norm-img}, which is averaged over multiple random seeds. We re-use the original code of \cite{geiping2020inverting}\footnote{\url{https://github.com/JonasGeiping/invertinggradients}} for attack. All experiments are implemented in PyTorch and conducted on a Linux workstation. All experimental configurations are detailed in Table \ref{table: Values_of_model_coefficient}.

\subsection{Results and Analysis}\label{sec:result}

\paragraph{\textbf{Better Privacy-Utility Trade-off}} Figure \ref{fig:eps-acc} shows the privacy-utility tradeoff curves under different privacy frameworks and defense mechanisms. We observe that \textbf{\textit{\underline{1)}}} \textcolor{black}{Our method consistently achieves better accuracy than baseline methods at fixed budgets.} \textbf{\textit{\underline{2)}}} \textcolor{black}{The utility gains are generally larger at more restrictive budgets.} As shown in Table \ref{tab:dp-eps-acc}, the performance gain increases from $1.36\%$ (\textcolor{black}{$\chi=800$}) to $8.48\%$ (\textcolor{black}{$\chi=200$}) on the MNIST dataset. \textbf{\textit{\underline{3)}}} Our method is more robust to extremely large distortion. When $\chi$ decreases further to $80$ (which means more noise is added to the model), the distortion noise becomes large enough to disrupt the training of the baseline method, which has $8.92\%$ accuracy and is worse than random guessing. Our method remains trainable and achieves $51.04\%$ accuracy. Thus, we answer \textbf{Q1} and \textbf{Q2}.

\paragraph{\textbf{Maintained Empirical Defense Effect}} As Table \ref{tab:emprical-privacy} shows, \textcolor{black}{our method and the baseline provide similar empirical protection against InvGrad at fixed budgets. The MSE and SSIM results support this comparison.} Figure \ref{fig:dlg-rec-img} shows that \textcolor{black}{InvGrad fails to reconstruct recognizable images for both methods, and the reconstructed images are visually similar}. Table \ref{tab:cap} further reports the empirical utility-privacy tradeoff against the \textcolor{black}{InvGrad attack} via the CAP metric. We observe that our method achieves a higher CAP value on all datasets and privacy frameworks.


\paragraph{\textbf{Intuitive Analysis of Distortion Learning}} Figure \ref{fig:delta-stat} shows the layer-wise distortion statistics to analyze the influence of the learning process. ``fc'' and ``conv'' denote fully-connected and convolutional layers, while ``w'' and ``b'' denote weights and biases. We observe that \textbf{\textit{\underline{1)}}} The distortion extent of some layers changes significantly, whereas that of other layers does not. This may intuitively explain the success of our algorithm in maintaining utility. For example, in the left-most figure of Figure \ref{fig:delta-stat}, the algorithm selectively amplifies(conv2.w, conv3.b), reduces(conv1.w, conv1.b, conv2.b, conv3.b) and maintains (fc.b, fc.w) the distortion extent of different layers to balance the privacy and utility. \textbf{\textit{\underline{2)}}} \textcolor{black}{The standard deviation in each layer changes only slightly across the tested settings, suggesting that the overall dispersion remains similar.}

\paragraph{\textbf{Additional Trade-off Results}} Table \ref{tab:pl-eps-acc} shows the detailed quantitative results of the privacy-utility trade-off under our privacy framework. Figure \ref{fig:clients} shows the client training process under a certain budget value, with faster and better convergence curves for our method. 
\paragraph{\textbf{Detailed Quantitative Defense Results}} \textcolor{black}{To assess whether the protection mechanisms achieve comparable empirical protection against the InvGrad attack, we quantitatively evaluate the MSE and SSIM between InvGrad-reconstructed images and the ground truth. Figure \ref{fig:dlg-rec-img} visualizes the defense effect at representative budgets, and Table \ref{tab:dp-mse} and Table \ref{tab:pl-mse} report MSE and SSIM results for all budgets. Across both tables, the SSIM values are at most $6.3\%$, while MSE is reported in the raw table units. The small mean differences across budget values indicate comparable empirical protection by our method and the baseline under this attack.}

{
\paragraph{\textbf{Reduced Step-wise Utility Loss}} Figure \ref{fig:stepwise-img} visualizes the step-wise utility loss (i.e., the signed utility difference $\epsilon_{u,t}^{(k)}$ caused by the current-round distortion) on the MNIST and FMNIST datasets. We observe that for the \textcolor{black}{LS-Static} baseline approach, the per-step utility loss is highly fluctuating and excessively large, which can contribute to greater global utility degradation through error accumulation. In contrast, our proposed \textcolor{black}{LS-Learn} method dynamically adapts the distortion by continuously optimizing the perturbation parameters at each training step. By doing so, our algorithm consistently maintains the step-wise utility loss at a significantly lower level compared to the baseline throughout the training process.

\begin{figure}[t]
    \centering
    \begin{overpic}[width=0.39\textwidth]{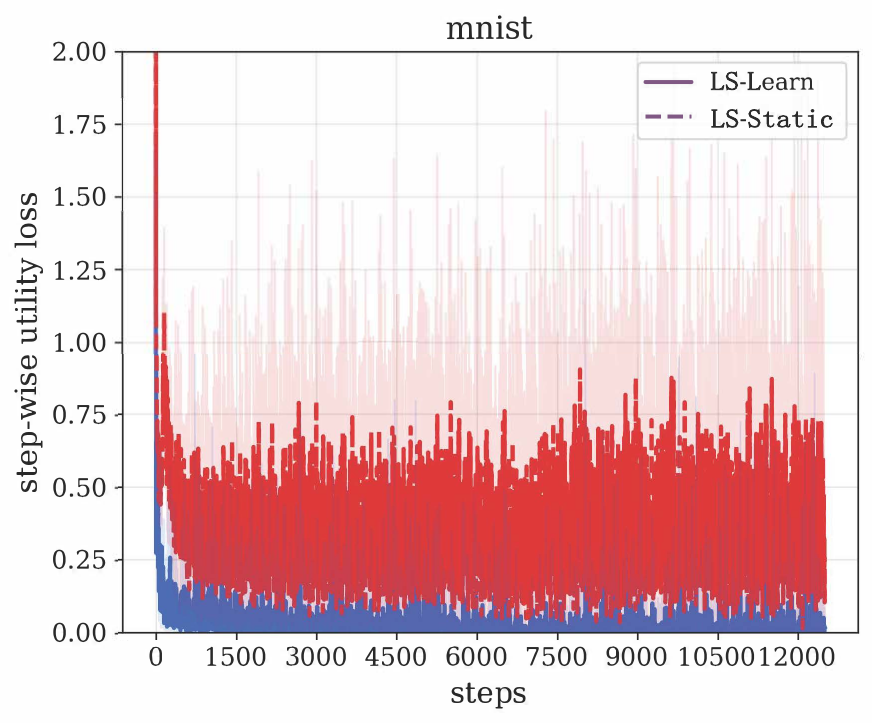}
    \end{overpic}
    \begin{overpic}[width=0.39\textwidth]{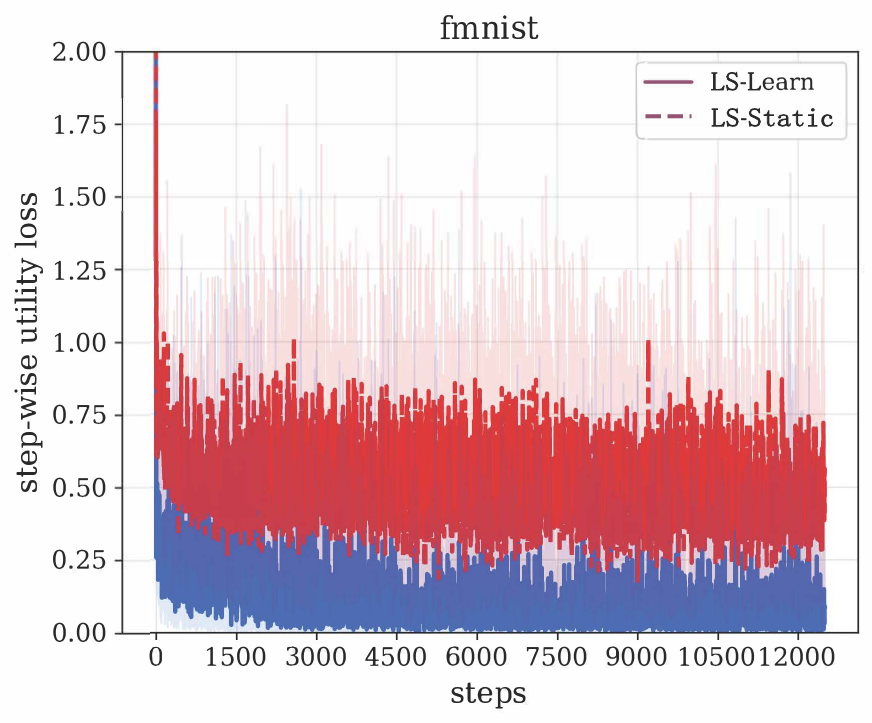}
    \end{overpic}
    \caption{The step-wise utility loss curves on MNIST and FMNIST datasets. By solving the per-round optimization problem, the proposed \textcolor{black}{LS-Learn} method consistently achieves a lower \textcolor{black}{current-round utility loss} compared to the \textcolor{black}{LS-Static} baseline.}
    \label{fig:stepwise-img}
\end{figure}

\paragraph{\textbf{Empirical Average Squared Gradient Norm}} Figure \ref{fig:average_squared_gradient_norm-img} shows the average squared gradient norm of \textcolor{black}{LS-Learn} during training under varying privacy budgets averaged over multiple random seeds. We observe that the empirical average squared gradient norm varies with the privacy budget. A larger budget ($\chi=1000$, implying less injected noise) is associated with a lower average squared gradient norm. In contrast, decreasing the budget to $\chi=800$ is associated with a higher value. Under a relatively strict budget of $\chi=600$, the average squared gradient norm fails to stabilize at a low level and instead rebounds significantly. \textcolor{black}{This trend is qualitatively consistent with \pref{thm: converge_rate_mt}, whose bound depends on the expected squared norm of the aggregate distortion rather than directly on the privacy budget.}

\begin{figure}[t]
    \centering
    \includegraphics[width=0.49\textwidth]{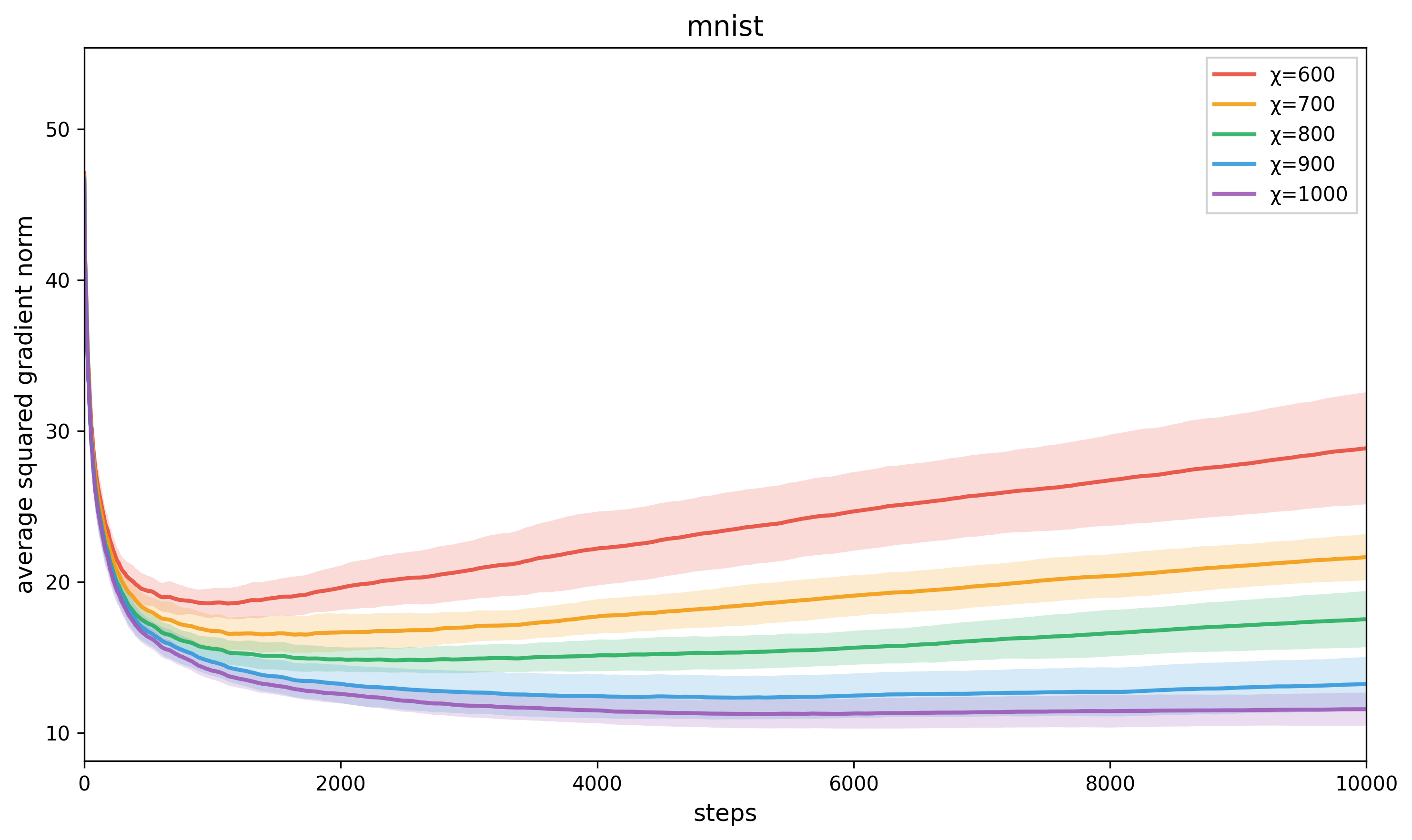}
    \includegraphics[width=0.49\textwidth]{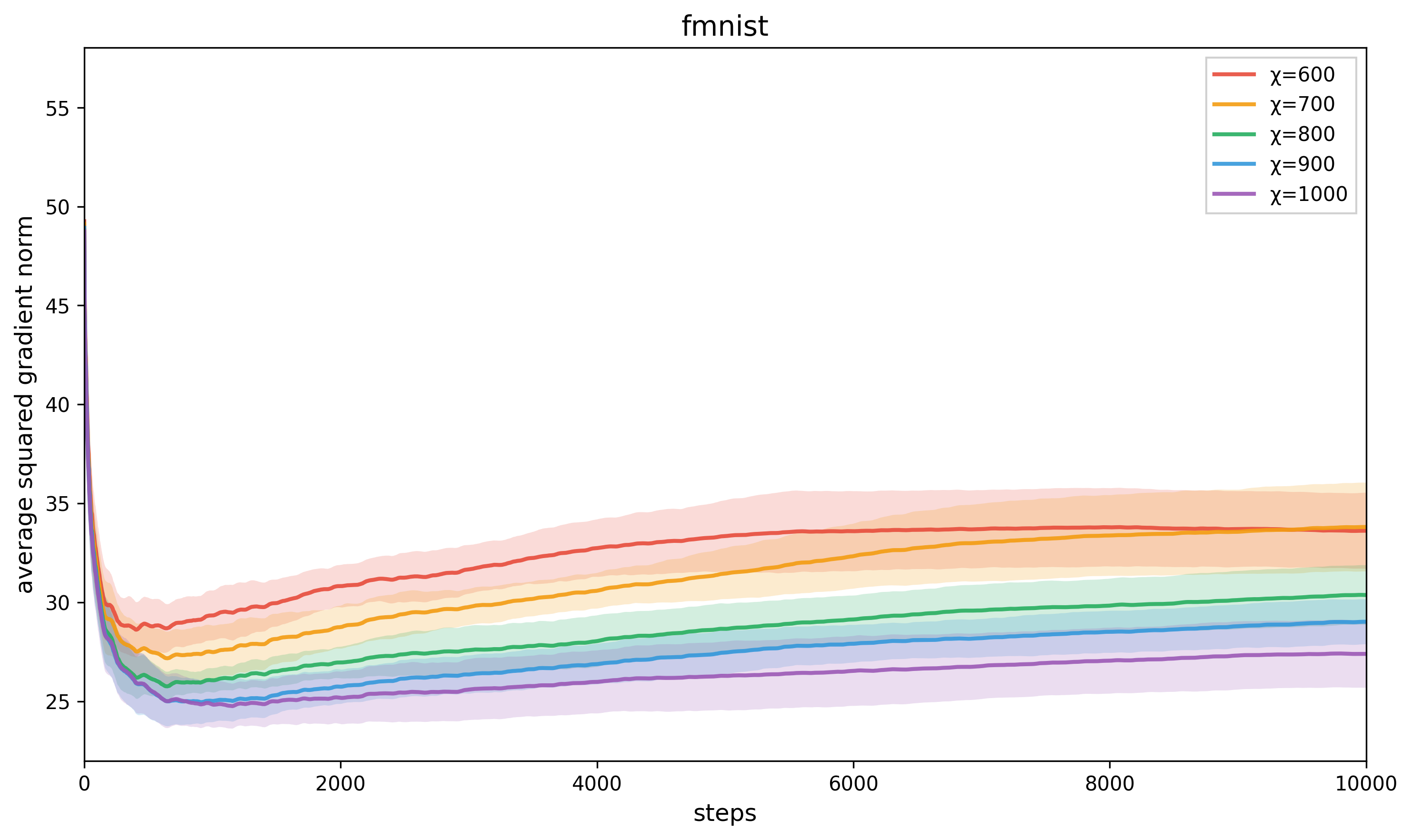}
    \caption{The average squared gradient norm of \textcolor{black}{LS-Learn} during training under varying privacy budgets $\chi$. \textcolor{black}{In these experiments, smaller budgets are associated with larger empirical values.}}
    \label{fig:average_squared_gradient_norm-img}
\end{figure}
}
\begin{table*}[t]
\centering
\caption{\textcolor{black}{The MSE$\uparrow$ (higher means larger reconstruction error and stronger empirical defense) and SSIM$\downarrow$ (lower means weaker structural similarity and stronger empirical defense) of reconstructed images under the Laplace-scale calibration setting. The SSIM values are at most $6.3\%$, while MSE is reported in the raw table units. The small method-wise differences indicate comparable empirical protection against InvGrad.}}
\begin{tabular}{@{}c|c|c|ccccc|c@{}}
\toprule
Dataset                 & Metric                & Method    & 80     & 200    & 400    & 600    & 800    & Diff Mean             \\ \midrule
\multirow{6}{*}{MNIST}  & \multirow{3}{*}{MSE}  & Identical & 3.48   & 2.61   & 2.44   & 2.22   & 2.23   &                       \\
                        &                       & Learn     & 3.30   & 2.48   & 2.43   & 2.51   & 2.28   &                       \\ \cmidrule(l){3-9} 
                        &                       & diff      & -0.18  & -0.13  & -0.01  & 0.28   & 0.04   & {\ul \textbf{0.00}}   \\ \cmidrule(l){2-9} 
                        & \multirow{3}{*}{SSIM} & Identical & 3.0\%  & 4.4\%  & 3.1\%  & 2.8\%  & 2.3\%  &                       \\
                        &                       & Learn     & 4.0\%  & 3.5\%  & 2.8\%  & 3.1\%  & 2.2\%  &                       \\ \cmidrule(l){3-9} 
                        &                       & diff      & 1.1\%  & -0.9\% & -0.3\% & 0.2\%  & -0.1\% & {\ul \textbf{0.0\%}}  \\ \midrule
\multirow{6}{*}{FMNIST} & \multirow{3}{*}{MSE}  & Identical & 2.09   & 1.96   & 2.01   & 1.91   & 1.88   &                       \\
                        &                       & Learn     & 2.05   & 1.95   & 1.92   & 1.86   & 1.84   &                       \\ \cmidrule(l){3-9} 
                        &                       & diff      & -0.04  & -0.01  & -0.09  & -0.04  & -0.04  & {\ul \textbf{-0.04}}  \\ \cmidrule(l){2-9} 
                        & \multirow{3}{*}{SSIM} & Identical & 6.3\%  & 2.7\%  & 2.7\%  & 2.3\%  & 2.0\%  &                       \\
                        &                       & Learn     & 5.2\%  & 2.6\%  & 1.2\%  & 1.7\%  & 2.5\%  &                       \\ \cmidrule(l){3-9} 
                        &                       & diff      & -1.1\% & -0.1\% & -1.5\% & -0.5\% & 0.5\%  & {\ul \textbf{-0.5\%}} \\ \bottomrule
\end{tabular}
\label{tab:dp-mse}
\end{table*}

\begin{table*}[t]
\centering
\caption{\textcolor{black}{The MSE$\uparrow$ and SSIM$\downarrow$ of InvGrad-reconstructed images under the distribution-independent privacy-leakage calibration setting. The SSIM values are at most $5.8\%$, while MSE is reported in the raw table units. The small method-wise differences indicate comparable empirical protection against InvGrad.}}
\label{tab:pl-mse}
\begin{tabular}{@{}c|c|c|ccccc|c@{}}
\toprule
Dataset                 & Metric                & Method    & 0.96   & 0.97   & 0.98   & 0.99  & 0.995  & Diff Mean             \\ \midrule
\multirow{6}{*}{MNIST}  & \multirow{3}{*}{MSE}  & Identical & 3.45   & 3.31   & 2.84   & 2.67  & 1.91   &                       \\
                        &                       & Learn     & 3.37   & 3.10   & 2.74   & 2.24  & 2.35   &                       \\ \cmidrule(l){3-9} 
                        &                       & diff      & -0.08  & -0.21  & -0.10  & -0.43 & 0.44   & {\ul \textbf{-0.08}}  \\ \cmidrule(l){2-9} 
                        & \multirow{3}{*}{SSIM} & Identical & 2.9\%  & 4.0\%  & 2.4\%  & 2.5\% & 3.3\%  &                       \\
                        &                       & Learn     & 2.4\%  & 3.3\%  & 2.3\%  & 2.7\% & 3.0\%  &                       \\ \cmidrule(l){3-9} 
                        &                       & diff      & -0.5\% & -0.7\% & -0.1\% & 0.2\% & -0.3\% & {\ul \textbf{-0.3\%}} \\ \midrule
\multirow{6}{*}{FMNIST} & \multirow{3}{*}{MSE}  & Identical & 2.10   & 2.04   & 1.99   & 1.89  & 1.68   &                       \\
                        &                       & Learn     & 2.09   & 2.07   & 1.91   & 1.87  & 1.84   &                       \\ \cmidrule(l){3-9} 
                        &                       & diff      & -0.01  & 0.03   & -0.08  & -0.03 & 0.16   & {\ul \textbf{0.02}}   \\ \cmidrule(l){2-9} 
                        & \multirow{3}{*}{SSIM} & Identical & 4.5\%  & 4.6\%  & 4.4\%  & 1.9\% & 2.7\%  &                       \\
                        &                       & Learn     & 5.5\%  & 5.8\%  & 2.1\%  & 1.9\% & 2.0\%  &                       \\ \cmidrule(l){3-9} 
                        &                       & diff      & 1.1\%  & 1.2\%  & -2.3\% & 0.0\% & -0.8\% & {\ul \textbf{-0.2\%}} \\ \bottomrule
\end{tabular}
\end{table*}

\begin{table*}[!ht]
\footnotesize
  \centering
  \setlength{\belowcaptionskip}{15pt}
  \caption{Experimental configuration.} 
  \label{table: Values_of_model_coefficient}
    \begin{tabular}{ccc}
    \toprule
    Category & Parameter & Value\cr
    \midrule
    Federated Learning Setup & Backbone Model & LeNetZhu\cr
    & Number of Clients & 4\cr
    & \textcolor{black}{Local-batch Update Steps} & \textcolor{black}{1}\cr
    & Optimizer & FedSGD\cr
    & Local Batch Size & \textcolor{black}{4}\cr
    \midrule
    Attack Setting & Attack Method & InvGrad\cr
    & Attack Iterations & 1600\cr
    & Attack Optimizer & Adam\cr
    & Attack Learning Rate & 1\cr
    & Total Variation Coefficient & 1e-5\cr
    & \textcolor{black}{Target (Local) Batch Size} & \textcolor{black}{4}\cr
    & Local Training Epoch & 1\cr
    \midrule
    Distortion Learning & Learning Iterations & 10 \cr
    & Distortion Learning Rate & 0.1\cr
    & \textcolor{black}{Distortion Optimizer} & \textcolor{black}{SGD}\cr
    & \textcolor{black}{Negative-Norm Coefficient $\lambda$} & \textcolor{black}{$10^{-5}$}\cr
    & Distortion Range & \textcolor{black}{$[l_t^{(k)},u_t^{(k)}]$}\cr
    & \textcolor{black}{Sensitivity/Gradient Clip Threshold} & 500\cr
    \bottomrule
    \end{tabular}
\end{table*}

\bibliography{main}
\bibliographystyle{ACM-Reference-Format}

\newpage

\onecolumn
\appendix
\renewcommand{\theassumption}{\Alph{section}.\arabic{assumption}}
\renewcommand{\thedefinition}{\Alph{section}.\arabic{definition}}
\renewcommand{\thethm}{\Alph{section}.\arabic{thm}}
\renewcommand{\thecor}{\Alph{section}.\arabic{cor}}
\renewcommand{\theexmp}{\Alph{section}.\arabic{exmp}}
\let\fednfloldthmhead\thmhead
\renewcommand{\thmhead}[3]{\fednfloldthmhead{#1}{\bluerev{#2}}{#3}}
\section{Broader Impact and Limitations}
We propose a general distortion learning framework for building protection mechanisms that can achieve a better trade-off between privacy and utility. Our work is expected to promote further research on utility-friendly protection mechanisms and thereby help address privacy-leakage threats in FL. In this paper, we mainly demonstrate the effectiveness of the overall framework \textcolor{black}{on image-classification tasks. Additional task-specific designs are needed to apply our framework to diverse FL scenarios; this is a limitation of \bluerev{this study} and a direction for future work.}

Regarding the practical implementation of our framework, the choice of hyperparameters for the distortion learning process (Algorithm 2), such as the learning rate $\gamma_t$ and the number of iterations $M$, is crucial for achieving the best results. We recommend that practitioners tune these parameters based on their specific application. The optimal configuration is inherently tied to the characteristics of the dataset, the complexity of the model, and the desired balance between privacy and utility.

Data complexity also plays a significant role in the practical deployment of our framework. Scenarios involving more complex data (e.g., high-resolution images) typically require models with a higher number of parameters. While our framework is scalable, the computational cost of the LearnDistortion procedure (Algorithm 2) increases with model size. \textcolor{black}{The number of optimization steps, $M$, serves as a hyperparameter that balances the stationarity quality of the learned nonconvex distortion trajectory against the per-round computational expense; in the convex exact-gradient regime, it also controls \bluerev{the $O(1/M)$ optimality-gap benchmark}.} \bluerev{Accordingly,} we can select the hyperparameter $M$ based on the complexity of the data. Specifically, more complex tasks may require a larger $M$, a choice contingent on available computational resources.

The proposed framework is inherently scalable due to its design. The most intensive new computation, the LearnDistortion procedure (Algorithm 2), is performed entirely on the client side. As clients operate in parallel, increasing the number of participating units ($K$) does not increase the per-client computational time. The central server's workload consists only of aggregating the distorted model parameters, a standard operation in federated learning whose overhead grows linearly with $K$. Therefore, our framework maintains the same high level of scalability as conventional federated learning algorithms like FedSGD. While our experiments demonstrate our method's effectiveness in a typical FL setting, a comprehensive empirical study on a very large-scale system with thousands of clients would be a valuable direction for future work to investigate the practical effects of network latency and communication bottlenecks.

\bluerev{The theoretical analysis presented in this work} primarily focuses on the trade-off between privacy and utility, and does not explicitly model the impact of communication constraints such as channel capacity. In practical deployments, communication bandwidth is a significant bottleneck that affects the overall efficiency (e.g., training time). A formal extension of the optimization problem to a three-way trade-off between privacy, utility, and communication efficiency represents an important avenue for future research.

{\color{black}
\section{Analysis for \pref{thm: distortion_stationarity_mt} and \pref{cor: convex_suboptimality_mt}}
\label{app: utility_loss_near_optimal}

This section first proves the average first-order stationarity guarantee for the nonconvex $L_2$-annular projected-SGD trajectory and then \bluerev{establishes the convex exact-gradient utility-loss corollary}.

\subsection{Analysis for \pref{thm: distortion_stationarity_mt}}

\begin{proof}
Fix $t$, $k$, and the realized mini-batch, and abbreviate $\phi=\phi_{t,\mathcal B}^{(k)}$, $\mathcal A=\mathcal A_t^{(k)}$, and $\gamma=\gamma_t$. Let
\begin{align}
d_m&\triangleq\alpha_{m+1}-\alpha_m,
&e_m&\triangleq G_m-\nabla\phi(\alpha_m),
&a&\triangleq\frac{1-L_\alpha\gamma}{2\gamma}>0.
\end{align}

{\color{black}
Set $z_m=\alpha_m-\gamma G_m$. Because $\alpha_{m+1}$ is a global Euclidean projection of $z_m$ onto the closed, possibly nonconvex set $\mathcal A$, while $\alpha_m\in\mathcal A$, projection-distance minimality gives
\begin{align}
\|\alpha_{m+1}-z_m\|_2^2
\le\|\alpha_m-z_m\|_2^2.
\end{align}
Substituting $d_m$ and $z_m$ and expanding both sides yields
\begin{align}
\langle G_m,d_m\rangle
\le-\frac{1}{2\gamma}\|d_m\|_2^2.
\label{eq: nonconvex_projection_descent}
\end{align}
This step uses neither convexity nor uniqueness of the selected projection.
}

Applying \pref{eq: restricted_smooth_descent}, using $\nabla\phi(\alpha_m)=G_m-e_m$, and then applying \pref{eq: nonconvex_projection_descent}, we obtain
\begin{align}
\phi(\alpha_{m+1})
&\le\phi(\alpha_m)
+\langle G_m,d_m\rangle
-\langle e_m,d_m\rangle
+\frac{L_\alpha}{2}\|d_m\|_2^2\\
&\le\phi(\alpha_m)-a\|d_m\|_2^2
-\langle e_m,d_m\rangle.
\end{align}
Young's inequality gives
\begin{align}
-\langle e_m,d_m\rangle
\le\frac a2\|d_m\|_2^2
+\frac{1}{2a}\|e_m\|_2^2,
\end{align}
and hence
\begin{align}
\frac a2\|d_m\|_2^2
\le\phi(\alpha_m)-\phi(\alpha_{m+1})
+\frac{1}{2a}\|e_m\|_2^2.
\label{eq: inexact_nonconvex_one_step}
\end{align}
Taking expectations, summing \pref{eq: inexact_nonconvex_one_step} for $m=1,\ldots,M$, and using $\phi(\alpha_{M+1})\ge\phi_{\inf}$ and \pref{eq: inner_gradient_mse} gives
\begin{align}
\frac a2\sum_{m=1}^M\E[\|d_m\|_2^2]
\le\Delta_\phi+\frac{M\nu_\alpha^2}{2a}.
\end{align}
Since $R_m=-d_m/\gamma$ and $a=(1-L_\alpha\gamma)/(2\gamma)$, rearranging proves \pref{eq: nonconvex_step_stationarity}.

It remains to connect the selected projected step to first-order stationarity. A global projection onto a closed set supplies a proximal normal, so
\begin{align}
z_m-\alpha_{m+1}
\in N_{\mathcal A}^{P}(\alpha_{m+1}).
\end{align}
The proximal normal cone is a cone and is contained in the limiting normal cone. Therefore
\begin{align}
n_{m+1}
&\triangleq\frac{z_m-\alpha_{m+1}}{\gamma}
=R_m-G_m
\in N_{\mathcal A}(\alpha_{m+1}).
\end{align}
Using this particular normal vector, \pref{eq: restricted_gradient_lipschitz}, and $e_m=G_m-\nabla\phi(\alpha_m)$, we have
\begin{align}
\mathfrak s(\alpha_{m+1})
&\le\|\nabla\phi(\alpha_{m+1})+n_{m+1}\|_2\\
&\le(1+L_\alpha\gamma)\|R_m\|_2+\|e_m\|_2.
\end{align}
The inequality $(x+y)^2\le2x^2+2y^2$, followed by averaging, \pref{eq: inner_gradient_mse}, and \pref{eq: nonconvex_step_stationarity}, proves \pref{eq: nonconvex_normal_stationarity}.

Finally, if $e_m=0$, the inequality preceding Young's inequality gives directly
\begin{align}
a\|d_m\|_2^2
\le\phi(\alpha_m)-\phi(\alpha_{m+1}).
\end{align}
Telescoping this sharper inequality proves \pref{eq: exact_nonconvex_step_stationarity}; \pref{eq: exact_nonconvex_normal_stationarity} then follows from $\mathfrak s(\alpha_{m+1})\le(1+L_\alpha\gamma)\|R_m\|_2$. This completes the proof.
\end{proof}

\subsection{Analysis for \pref{cor: convex_suboptimality_mt}}

\begin{proof}
For the convex corollary, abbreviate $f=f_t^{(k)}$, $\mathcal C=\mathcal C_t^{(k)}$, $\alpha^*=\alpha_t^{(k)*}$, and $\gamma=\gamma_t$. The exact update is
\begin{align}
\alpha_{m+1}
=\operatorname{Proj}_{\mathcal C}
\left(\alpha_m-\gamma\nabla f(\alpha_m)\right).
\end{align}
Because $\mathcal C$ is closed and convex, its projection variational inequality gives, for every $\alpha\in\mathcal C$,
\begin{align}
\left\langle\nabla f(\alpha_m),\alpha_{m+1}-\alpha\right\rangle
\le\frac1\gamma
\left\langle\alpha_m-\alpha_{m+1},\alpha_{m+1}-\alpha\right\rangle.
\label{eq: projection_optimality_new}
\end{align}
Convexity and $L_\alpha$-smoothness of $f$, together with $\gamma\le1/L_\alpha$, imply
\begin{align}
f(\alpha_{m+1})-f(\alpha^*)
&\le
\left\langle\nabla f(\alpha_m),\alpha_{m+1}-\alpha^*\right\rangle
+\frac{1}{2\gamma}\|\alpha_{m+1}-\alpha_m\|_2^2\\
&\le\frac{1}{2\gamma}
\left(
\|\alpha_m-\alpha^*\|_2^2
-\|\alpha_{m+1}-\alpha^*\|_2^2
\right).
\label{eq: pgd_telescoping}
\end{align}
Applying \pref{eq: projection_optimality_new} with $\alpha=\alpha_m$ and using smoothness also shows that $f(\alpha_{m+1})\le f(\alpha_m)$. Therefore, summing \pref{eq: pgd_telescoping} over $m=1,\ldots,M$ gives
\begin{align}
M\bigl(f(\alpha_{M+1})-f(\alpha^*)\bigr)
&\le\sum_{m=1}^M
\bigl(f(\alpha_{m+1})-f(\alpha^*)\bigr)\\
&\le\frac{\|\alpha_1-\alpha^*\|_2^2}{2\gamma}.
\end{align}
Since $f=\epsilon_{u,t}^{(k)}$ and both $\alpha_1$ and $\alpha^*$ belong to $\mathcal C$, the diameter bound gives \pref{eq: per_round_suboptimality}.
\end{proof}
}

\section{Analysis for \pref{thm: converge_rate_mt}}
\label{app: converge_rate}

In this section, we analyze the average-stationarity guarantee of our proposed algorithm. The averaged stochastic gradient at round $t$ is denoted by $g_t$.

\begin{lem}\label{lem: the_gap_of_f_W_for_adjancent_w_mt}
\yellow{Let \pref{assump: global_objective} and \pref{assump: stochastic_gradient_distortion} hold, and choose $0<\eta\le1/(4L_F)$. Then
\begin{align}
\frac{\eta}{2}\E[\|\nabla F(W_{t-1})\|^2]
&\le \E[F(W_{t-1})-F(W_t)]
+L_F\eta^2\sigma_g^2\nonumber\\
&\quad+\left(\frac{1}{\eta}+L_F\right)
\E[\|\Delta_t\|^2].\label{eq: one_step_stationarity}
\end{align}}
\end{lem}

With the above lemma, we are ready to prove \pref{thm: converge_rate_mt}.

\subsection{Analysis for \bluerev{\pref{lem: the_gap_of_f_W_for_adjancent_w_mt}}}\label{app: the_gap_of_f_W_for_adjancent_w}

\begin{proof}
Let $q_t=\nabla F(W_{t-1})$. By the $L_F$-smoothness of $F$ and the update $W_t-W_{t-1}=-\eta g_t+\Delta_t$,
\yellow{\begin{align}
F(W_t)
&\le F(W_{t-1})
-\eta\langle q_t,g_t\rangle
+\langle q_t,\Delta_t\rangle
+\frac{L_F}{2}\|-\eta g_t+\Delta_t\|^2\\
&\le F(W_{t-1})
-\eta\langle q_t,g_t\rangle
+\langle q_t,\Delta_t\rangle
+L_F\eta^2\|g_t\|^2
+L_F\|\Delta_t\|^2.
\label{eq: perturbed_descent_step}
\end{align}}
\textcolor{black}{Here, $\mathcal F_{t-1}$ denotes the history $\sigma$-algebra containing $W_0$ and all randomness revealed through round $t-1$. Hence $W_{t-1}$ and $q_t=\nabla F(W_{t-1})$ are $\mathcal F_{t-1}$-measurable, while $g_t$ retains the stochasticity introduced at round $t$.}
{\color{black}
The unbiasedness and bounded-variance conditions imply
\yellow{\begin{align}
\E[\langle q_t,g_t\rangle\mid\mathcal F_{t-1}]&=\|q_t\|^2,\\
\E[\|g_t\|^2\mid\mathcal F_{t-1}]&\le\|q_t\|^2+\sigma_g^2.
\end{align}}
No zero-mean assumption is imposed on $\Delta_t$. Instead, Young's inequality gives
\yellow{\begin{align}
\langle q_t,\Delta_t\rangle
\le\frac{\eta}{4}\|q_t\|^2+\frac{1}{\eta}\|\Delta_t\|^2.
\end{align}}
Taking the conditional expectation of \pref{eq: perturbed_descent_step} and using $\eta\le1/(4L_F)$, we obtain
\yellow{\begin{align}
\E[F(W_t)\mid\mathcal F_{t-1}]
&\le F(W_{t-1})
-\frac{\eta}{2}\|q_t\|^2
+L_F\eta^2\sigma_g^2\\
&\quad+\left(\frac{1}{\eta}+L_F\right)
\E[\|\Delta_t\|^2\mid\mathcal F_{t-1}].
\end{align}}
Taking total expectations and rearranging proves \pref{eq: one_step_stationarity}.
}
\end{proof}

\subsection{Analysis for \pref{thm: converge_rate_mt}}

\begin{proof}
From \bluerev{\pref{lem: the_gap_of_f_W_for_adjancent_w_mt}}, summing over $t=1,\ldots,T$ gives
\yellow{\begin{align}
\frac{\eta}{2}\sum_{t=1}^T\E[\|\nabla F(W_{t-1})\|^2]
&\le F(W_0)-\E[F(W_T)]
+TL_F\eta^2\sigma_g^2\\
&\quad+\left(\frac{1}{\eta}+L_F\right)
\sum_{t=1}^T\E[\|\Delta_t\|^2].
\end{align}}
Dividing both sides by $\eta T/2$ proves \pref{eq: converge_rate_mt}. Since $F(W_T)\ge F_*$, replacing $\E[F(W_T)]$ by $F_*$ gives the stated upper bound on the transient term. If $\E[\|\Delta_t\|^2]\le\rho^2$ for every $t$, taking the limit superior gives the stationary-neighborhood bound.
\end{proof}

\section{Measurements for Privacy Leakage and Optional Distortion Set}\label{sec: discuss_measurements_for_privacy}
We first introduce the definition of privacy leakage. The privacy leakage is measured using the gap between the estimated dataset and the original dataset. The semi-honest attacker uses an optimization algorithm \cite{zhu2020deep, geiping2020inverting, zhao2020idlg, yin2021see} to reconstruct the original dataset of the client given the exposed model information $W$.

\begin{definition}[Privacy Leakage]
Let $\breve d^{(m)}$ represent the original $m$-th data, and let $d_i^{(m)}$ represent the $m$-th data inferred by the attacker at iteration $i$. Let $I$ represent the total number of attack iterations. The privacy leakage $\epsilon_p$ is defined as

\begin{equation}\label{eq: defi_privacy_leakage}
\epsilon_p=\left\{
\begin{array}{cl}
\frac{D - \frac{1}{I}\sum_{i = 1}^{I} \frac{1}{|\calD|}\sum_{m = 1}^{|\calD|}||d_i^{(m)} - \breve d^{(m)}||}{D}, &  I > 0\\
0,  &  I = 0\\ 
\end{array} \right.
\end{equation}
\textbf{Remark:} We assume that $||d_i^{(m)} - \breve d^{(m)}||\in [0,D]$, where $D$ is a positive constant. Therefore, $\epsilon_p\in [0,1]$.
\end{definition}

The precise definition for privacy leakage is introduced in \pref{eq: defi_privacy_leakage}, which is measured using the distance between the recovered data and the true data. However, it would be difficult to solve the optimization problem (\pref{eq: constraint_optimization_problem_ul}) using this definition directly. Instead, we derive bounds on privacy leakage using the scalar distortion intensity $\beta^{(k)}$.


{\color{black}
{\color{black}
\textcolor{black}{In this appendix, the superscript $(k)$ consistently identifies the client and the subscript $t$ identifies the communication round. In Lemma~\ref{lem: bound_for_privacy_leakage}, $t$ is fixed and is suppressed only in the abbreviated variables introduced there.}
To derive an attack-specific upper bound, fix a communication round $t$ and client $k$, condition on the received model $W_{t-1}$, and let $\breve{\calB}_t^{(k)}=\{\breve d^{(m)}\}_{m=1}^{B_t^{(k)}}$ be the actual data collection used to compute the undistorted local update, where $B_t^{(k)}:=|\breve{\calB}_t^{(k)}|$. Define the per-example and aggregate exposed-update maps by
\begin{align}
g_t^{(k)}(d)&:=W_{t-1}-\eta\nabla_W\calL(W_{t-1},d),\\
\overline g_t^{(k)}(\calD)&:=\frac{1}{|\calD|}\sum_{d\in\calD}g_t^{(k)}(d).
\end{align}
Because the local loss is the empirical average over $\breve{\calB}_t^{(k)}$, the FedSGD update gives
\begin{align}\label{eq: actual_update_identity}
\overline g_t^{(k)}(\breve{\calB}_t^{(k)})=\breve W_t^{(k)}.
\end{align}
The individual samples $\breve d^{(m)}$ and $d_i^{(m)}$ below belong to the fixed client $k$, so their client index is suppressed. The indices $t$ and $(k)$ are retained on the exposed-update maps to distinguish $g_t^{(k)}$ and $\overline g_t^{(k)}$ from the budget-to-intensity mapping $g^{(k)}$ used in the main text.
}

\begin{assumption}[Bi-Lipschitz Condition]\label{assump: two_sided_Lipschitz}
For all original and reconstructed examples under consideration, there exist constants $0<c_a\le c_b$ such that
\begin{align}
c_a\|g_t^{(k)}(d_1)-g_t^{(k)}(d_2)\|_2
\le \|d_1-d_2\|_2
\le c_b\|g_t^{(k)}(d_1)-g_t^{(k)}(d_2)\|_2.
\end{align}
\end{assumption}

\bluerev{For the actually released local update $W_t^{(k)}$ and the attacker's reconstruction $\calB_i^{(k)}=\{d_i^{(m)}\}_{m=1}^{B_t^{(k)}}$ at attack iteration $i$, define the aggregate matching residual}
{\color{black}
\begin{align}\label{eq: aggregate_matching_residual}
r_i^{(k)}(W_t^{(k)}):=\|\overline g_t^{(k)}(\calB_i^{(k)})-W_t^{(k)}\|_2.
\end{align}
}

\begin{assumption}[Sublinear Aggregate Matching Residual]\label{assump: matching_residual}
\bluerev{For the fixed round, client, actual update batch, attack protocol, and selected attack horizon $I\ge1$, assume that there exist $c_2>0$ and $p\in(0,1)$ such that}
{\color{black}
\begin{align}\label{eq: matching_residual_bound}
\sum_{i=1}^{I}r_i^{(k)}(W_t^{(k)})\le c_2I^p
\qquad\text{almost surely}.
\end{align}
}
When one common leakage score is used over an optimization feasible set, the same $c_2$ and $p$ are assumed to hold uniformly over the protected outputs in that set. If multiple attack horizons are considered, the same constants are additionally assumed over those horizons. For a pointwise application of the lemma below, only the realized output $W_t^{(k)}$ at the selected horizon is required to satisfy \pref{eq: matching_residual_bound}. \bluerev{This condition is stated directly in terms of the attacker's matching objective and the actual release; it does not posit a dataset whose aggregate exposed update is exactly $W_t^{(k)}$.}
\end{assumption}

\begin{lem}\label{lem: bound_for_privacy_leakage}
Let $I\ge1$, abbreviate $\breve W^{(k)}=\breve W_t^{(k)}$, and let the released update be
\begin{align}
W^{(k)}=\breve W^{(k)}+\alpha^{(k)},
\qquad
\beta^{(k)}:=\|\alpha^{(k)}\|_2=\|W^{(k)}-\breve W^{(k)}\|_2.
\end{align}
\bluerev{Under \bluerev{Assumptions~\ref{assump: two_sided_Lipschitz} and \ref{assump: matching_residual}}, if}
{\color{black}
\begin{align}
\beta^{(k)}\ge\frac{2c_2c_b}{c_a}I^{p-1},
\end{align}
}
then the reconstruction-based privacy leakage in \pref{eq: defi_privacy_leakage} satisfies, almost surely,
{\color{black}
\begin{align}\label{eq: direct_privacy_upper_bound}
\epsilon_p^{(k)}
\le 1-\frac{c_a\beta^{(k)}+c_2c_bI^{p-1}}{4D}.
\end{align}
}
\end{lem}

Accordingly, for a fixed attack protocol and horizon $I$, we use the right-hand side of \pref{eq: direct_privacy_upper_bound} as the attack-specific leakage score
\begin{align*}
\epsilon_p^{(k)}(\beta^{(k)})
:=1-\frac{c_a\beta^{(k)}+c_2c_bI^{p-1}}{4D}.
\end{align*}
This score is an upper-bound surrogate under the stated conditions, not a mechanism-independent privacy guarantee; on configurations satisfying the lemma, the proof together with Definition~\ref{eq: defi_privacy_leakage} also implies that it lies in $[0,1]$.

\subsection{Analysis for \bluerev{\pref{lem: bound_for_privacy_leakage}}}

\begin{proof}
For each attack iteration $i$, the lower side of the bi-Lipschitz condition and Jensen's (equivalently, the triangle) inequality for the average give
\begin{align}
\frac{1}{B_t^{(k)}}\sum_{m=1}^{B_t^{(k)}}
\|d_i^{(m)}-\breve d^{(m)}\|_2
&\ge \frac{c_a}{B_t^{(k)}}\sum_{m=1}^{B_t^{(k)}}
 \|g_t^{(k)}(d_i^{(m)})-g_t^{(k)}(\breve d^{(m)})\|_2\\
&\ge c_a\left\|\overline g_t^{(k)}(\calB_i^{(k)})-\overline g_t^{(k)}(\breve{\calB}_t^{(k)})\right\|_2.
\end{align}
{\color{black}
Using \pref{eq: actual_update_identity}, $W^{(k)}=\breve W^{(k)}+\alpha^{(k)}$, and the reverse triangle inequality, we further obtain
\begin{align}
c_a\left\|\overline g_t^{(k)}(\calB_i^{(k)})-\overline g_t^{(k)}(\breve{\calB}_t^{(k)})\right\|_2
&=c_a\|\overline g_t^{(k)}(\calB_i^{(k)})-\breve W^{(k)}\|_2\\
&\ge c_a\bigl(\|W^{(k)}-\breve W^{(k)}\|_2-\|\overline g_t^{(k)}(\calB_i^{(k)})-W^{(k)}\|_2\bigr)\\
&=c_a\beta^{(k)}-c_ar_i^{(k)}(W^{(k)})\\
&\ge c_a\beta^{(k)}-c_br_i^{(k)}(W^{(k)}),
\end{align}
where the last inequality uses $r_i^{(k)}(W^{(k)})\ge0$ and $c_a\le c_b$.
}

Averaging over $I$ iterations and applying \pref{eq: matching_residual_bound} yields
\begin{align}
D(1-\epsilon_p^{(k)})
&=\frac{1}{I}\sum_{i=1}^{I}\frac{1}{B_t^{(k)}}\sum_{m=1}^{B_t^{(k)}}
\|d_i^{(m)}-\breve d^{(m)}\|_2\\
&\ge c_a\beta^{(k)}-\frac{c_b}{I}\sum_{i=1}^{I}r_i^{(k)}(W^{(k)})\\
&\ge c_a\beta^{(k)}-c_2c_bI^{p-1}.
\end{align}
Set $A=c_a\beta^{(k)}$ and $B=c_2c_bI^{p-1}$. The stated lower bound on $\beta^{(k)}$ implies $A\ge2B$, and hence
\begin{align}
A-B\ge\frac{A}{2}\ge\frac{A+B}{4}.
\end{align}
Consequently,
\begin{align}
D(1-\epsilon_p^{(k)})
\ge\frac{c_a\beta^{(k)}+c_2c_bI^{p-1}}{4},
\end{align}
which is equivalent to \pref{eq: direct_privacy_upper_bound}.
\end{proof}
}

\end{document}